\documentclass[letterpaper]{article} % DO NOT CHANGE THIS
\usepackage{aaai24}  % DO NOT CHANGE THIS
\usepackage{times}  % DO NOT CHANGE THIS
\usepackage{helvet}  % DO NOT CHANGE THIS
\usepackage{courier}  % DO NOT CHANGE THIS
\usepackage[hyphens]{url}  % DO NOT CHANGE THIS
\usepackage{graphicx} % DO NOT CHANGE THIS
\usepackage{natbib}  % DO NOT CHANGE THIS AND DO NOT ADD ANY OPTIONS TO IT
\usepackage{caption} % DO NOT CHANGE THIS AND DO NOT ADD ANY OPTIONS TO IT
\usepackage{algorithm}
\usepackage{algorithmic}

\usepackage{newfloat}
\usepackage{listings}
\DeclareCaptionStyle{ruled}{labelfont=normalfont,labelsep=colon,strut=off} % DO NOT CHANGE THIS
\floatstyle{ruled}
\newfloat{listing}{tb}{lst}{}
\floatname{listing}{Listing}
\usepackage{multirow}
\usepackage{booktabs}
\usepackage{tabularx}
\usepackage{amssymb}
\usepackage{amsmath}
\usepackage{enumitem}
\usepackage{colortbl}
\usepackage{xcolor}
\usepackage{rotating}
\usepackage{pdflscape}
\usepackage{seqsplit}
\usepackage{longtable}

\newcommand{\MHA}{\mathrm{MHA}}
\newcommand{\Attn}{\mathrm{Attention}}
\newcommand{\FFN}{\mathrm{FFN}}
\newcommand{\LN}{\mathrm{LN}}

\title{MolTailor: Tailoring Chemical Molecular Representation to Specific Tasks via Text Prompts}
\author{
    Haoqiang Guo,
    Sendong Zhao\thanks{Corresponding author},
    Haochun Wang,
    Yanrui Du,
    Bing Qin
}
\affiliations{
    Research Center for Social Computing and Information Retrieval, Harbin Institute of Technology, China\\
    \{hqguo, sdzhao, hchwang, yrdu, bqing\}@ir.hit.edu.cn
}

\usepackage{bibentry}
\begin{document}

\maketitle

\begin{abstract}
Deep learning is now widely used in drug discovery, providing significant acceleration and cost reduction. As the most fundamental building block, molecular representation is essential for predicting molecular properties to enable various downstream applications. Most existing methods attempt to incorporate more information to learn better representations. However, not all features are equally important for a specific task. Ignoring this would potentially compromise the training efficiency and predictive accuracy. To address this issue, we propose a novel approach, which treats language models as an agent and molecular pretraining models as a knowledge base. The agent accentuates task-relevant features in the molecular representation by understanding the natural language description of the task, just as a tailor customizes clothes for clients. Thus, we call this approach \textbf{MolTailor}. Evaluations demonstrate MolTailor's superior performance over baselines, validating the efficacy of enhancing relevance for molecular representation learning. This illustrates the potential of language model guided optimization to better exploit and unleash the capabilities of existing powerful molecular representation methods. Our code is available at \url{https://github.com/SCIR-HI/MolTailor}.
\end{abstract}

\section{Introduction}
In recent years, AI technology has been widely applied to various stages of drug design, such as compound synthesis and screening, etc. \cite{mamoshina2016applications}. This has greatly improved the efficiency and reduced the cost of drug development. Molecular representation learning serves as the cornerstone for AI techniques in drug design, empowering a wide range of downstream tasks such as molecular property prediction and drug-drug interaction judgment \cite{xia2023systematic}. Molecular representations are essentially vector embeddings for molecules, analogous to word embeddings in NLP. The idea of encoding molecules as mathematical vectors dates back to the 1940s \cite{wiener1947structural}.

The development of molecular representation learning can be roughly divided into four phases. In the first phase, molecular representations were constructed through expert knowledge, which can be categorized into two types: descriptors \cite{wiener1947structural} and fingerprints \cite{rogers2010extended}. Descriptors focus on the inherent chemical characteristics of molecules, while fingerprints contain structural information. In the second phase, inspired by deep learning, some works \cite{gilmer2017neural, yang2019analyzing, song2020communicative} started to learn molecular representations from labeled data using supervised learning. However, the scarcity of labeled data, due to the expensive experimental cost and time required, limits further improvements to model performance and generalization.

\begin{figure}[t]
    \centering
    \includegraphics[width=1\columnwidth]{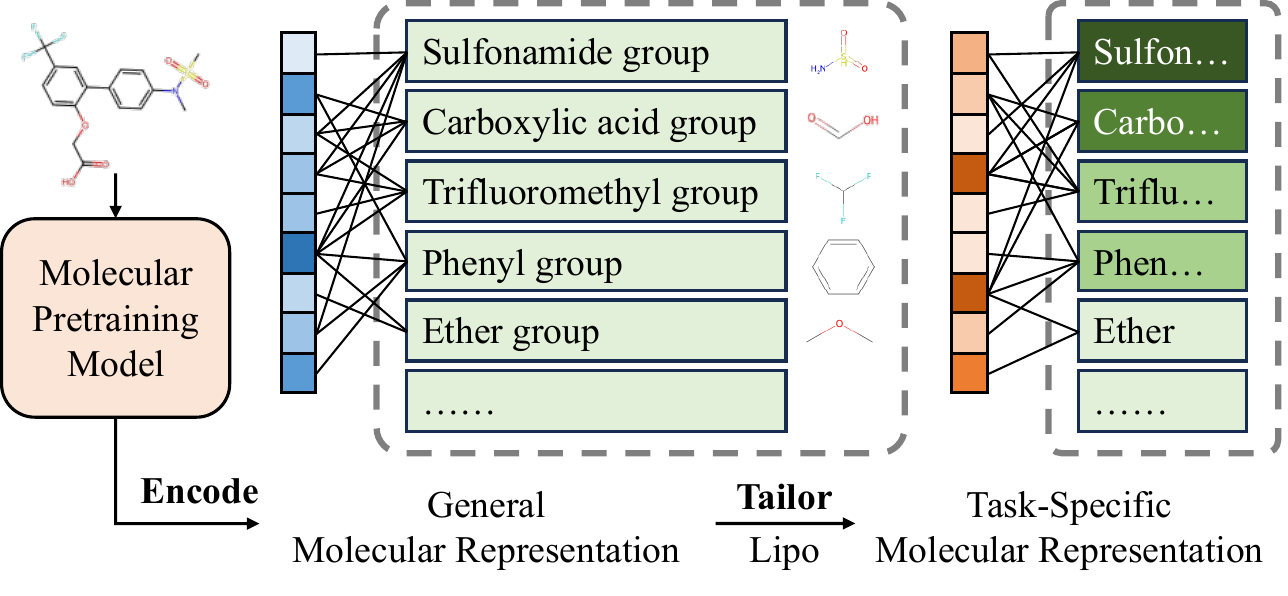} % Reduce the figure size so that it is slightly narrower than the column. Don't use precise values for figure width.This setup will avoid overfull boxes.
    \caption{Most existing molecular pretraining models (e.g. Grover, MolCLR) attempt to encode as much molecular information as possible (e.g. various functional groups and molecular weight) into a vector to obtain general molecular representation. However, for specific downstream tasks (e.g. Lipo, predicting lipophilicity of compounds), features are not equally important (e.g. Sulfonamide and Carboxylic acid groups significantly increase the hydrophilicity, being more critical than the remaining groups). By understanding task descriptions, MolTailor adjusts the weights of different features in the representation to obtain task-specific molecular representation.}
    \label{fig:motivation}
\end{figure}

The success of pretrained language models in NLP \cite{devlin2018bert} has demonstrated the potential of self-supervised learning. Building upon this, in the third phase, some works pretrained sequence-based models to learn molecular representation by mimicking successful NLP pretrained language models. \cite{chithrananda2020chemberta, kim2021merged, ross2022large}. Concurrently, others pretrained graph-based models \cite{hu2019strategies, you2020graph}. More recently, in the fourth phase, researchers are no longer limited to pretraining solely on the molecular data itself. Instead, they are attempting to incorporate additional information, such as knowledge graphs \cite{fang2022molecular} and textual information \cite{zeng2022deep, edwards2022translation, su2022molecular,chen2023artificially}. Notably, molecule-text multimodal learning has received increasing attention recently and achieved promising results.

As introduced above, most existing works merely strive to inject more information into the representations without utilizing task information as prior knowledge, which could compromise the training efficiency, as shown in Fig. \ref{fig:motivation}. Inspired by this observation, we propose a new method called \textbf{MolTailor}, where a language model acts as a tailor, adapting general molecular representations (ready-made clothing) into task-specific ones (customized clothing) based on user needs.

To achieve this goal, we adopt a dual-tower model structure with one language pretraining module and one molecular pretraining module, joint together through a cross-attention module. Meanwhile, we constructed a new pretraining task: \textbf{M}olecule-\textbf{T}ext \textbf{M}ulti-\textbf{T}ask \textbf{R}egression (\textbf{MT-MTR}). The dataset for this task consists of (molecule, task description, regression labels) triplets. Here, the molecule is represented as a SMILES string \cite{weininger1988smiles}, the task description is a text prompt describing molecular properties that are most helpful for solving the task, and the regression labels are values of properties mentioned in the task description. The model needs to predict the regression labels based on the SMILES and task description. This pretraining task teaches the model to enhance the weights of task-relevant properties in the molecular representation according to the task description. Our contributions can be summarized as follows:

\begin{itemize}
    \item We propose MolTailor, the first approach to generate task-specific molecular representations via text prompts, which provides a new perspective on text-molecule multimodal learning: not only injecting the knowledge in texts into molecular representations but also utilizing the reasoning capacity of language models.
    \item We construct MT-MTR, a new molecule-text multimodal pretraining task. This task teaches the model the capabilities of instruction following and adapting molecular representations.
    \item We comprehensively evaluate across eight subsets of the MoleculeNet, thereby demonstrating the effectiveness of task-specific molecular representation learning.
\end{itemize}

\section{Related Work}
\subsubsection{Molecular Pretraining Models.}
Models can be roughly categorized into the following three types: sequence-based, graph-based, and external knowledge \cite{xia2023systematic}. \textbf{1. Sequence-based}: Representing molecules as SMILES strings or other sequences and using language models as backbones for pretraining. SMILES-BERT \cite{wang2019smiles} and ChemBERTa \cite{chithrananda2020chemberta} use Masked Language Modeling (MLM) for pretraining tasks, while CHEM-BERT \cite{kim2021merged}, ChemBERTa-2 \cite{ahmad2022chemberta} additionally incorporate property prediction tasks. MM-Deacon \cite{guo2021multilingual} uses contrastive learning with SMILES and INPAC as parallel inputs. \textbf{2. Graph-based}: Representing molecules as graphs and using graph neural networks for pretraining. GROVER \cite{rong2020self} and Mole-BERT \cite{xia2022mole} use Mask Component Modeling (MCM) for pretraining, while GraphCL \cite{you2020graph} and MolCLR \cite{wang2022molecular} employ contrastive learning. Denoising \cite{zaidi2022pre} and GeoSSL \cite{liu2022molecular} are trained with denoising methods inspired by denoising diffusion models. \textbf{3. External knowledge}: GraphMVP \cite{liu2021pre} and 3D Infomax \cite{stark20223d} incorporate 3D structures as supplementary information. KCL \cite{fang2022molecular} uses knowledge graphs to enhance molecular representations.

\subsubsection{Image-Text Multimodal Pretraining Models.}
Models can be divided into dual-tower and single-tower structures based on architectures: \textbf{1. Dual-tower structure}: ViLBERT \cite{lu2019vilbert} introduces the co-attentional transformer (Co-TRM) layer, which effectively integrates information from both modalities. CLIP \cite{radford2021learning} adopts contrastive learning for pretraining and achieves significant performance gains. The CoCa \cite{yu2022coca} model uses dual towers of an image encoder and a text decoder. \textbf{2. Single-tower structure}: VisualBERT \cite{li2019visualbert} takes texts and images as a unified input for learning. ViLT \cite{kim2021vilt} adopts a patch projection approach, greatly improving the processing speed. BeiT-3 \cite{wang2023image} uses distinct Feedforward Neural Network (FFN) layers for different modalities while sharing attention modules.

\subsubsection{Molecule-Text Multimodal Pretraining Models.} 
Similarly, models can be divided into dual-tower and single-tower structures based on architectures: \textbf{1. Single-tower structure}: Using a language model as the backbone for further pretraining. KV-PLM \cite{zeng2022deep} and MolXPT \cite{liu2023molxpt} inject SMILES into literature text by locating molecule names to obtain mixed corpora and conduct MLM and language pretraining (LM) respectively. MolT5 \cite{edwards2022translation} does replace corrupted spans pretraining on molecular and text data, meanwhile, uses mutual translation between molecules and textual descriptions as downstream tasks. Text+Chem T5 \cite{christofidellis2023unifying} and ChatMol \cite{zeng2023interactive} use multi-task learning for pretraining. GIMLET \cite{zhao2023gimlet} takes graphs and texts as a unified input and uses instruction-based supervised property prediction for pretraining. \textbf{2. Dual-tower structure}: MoMu \cite{su2022molecular} and MoleculeSTM \cite{liu2022multi} do contrastive learning on molecule-description pairs. CLAMP \cite{seidl2023enhancing} does contrastive learning on molecule-bioassay pairs. Additionally, ChemCrow \cite{bran2023augmenting} enhances large language models (LLMs) with chemical tools to accomplish real-world chemical tasks.

\section{Methodology}

\begin{figure*}[t]
    \centering
    \includegraphics[width=1\textwidth]{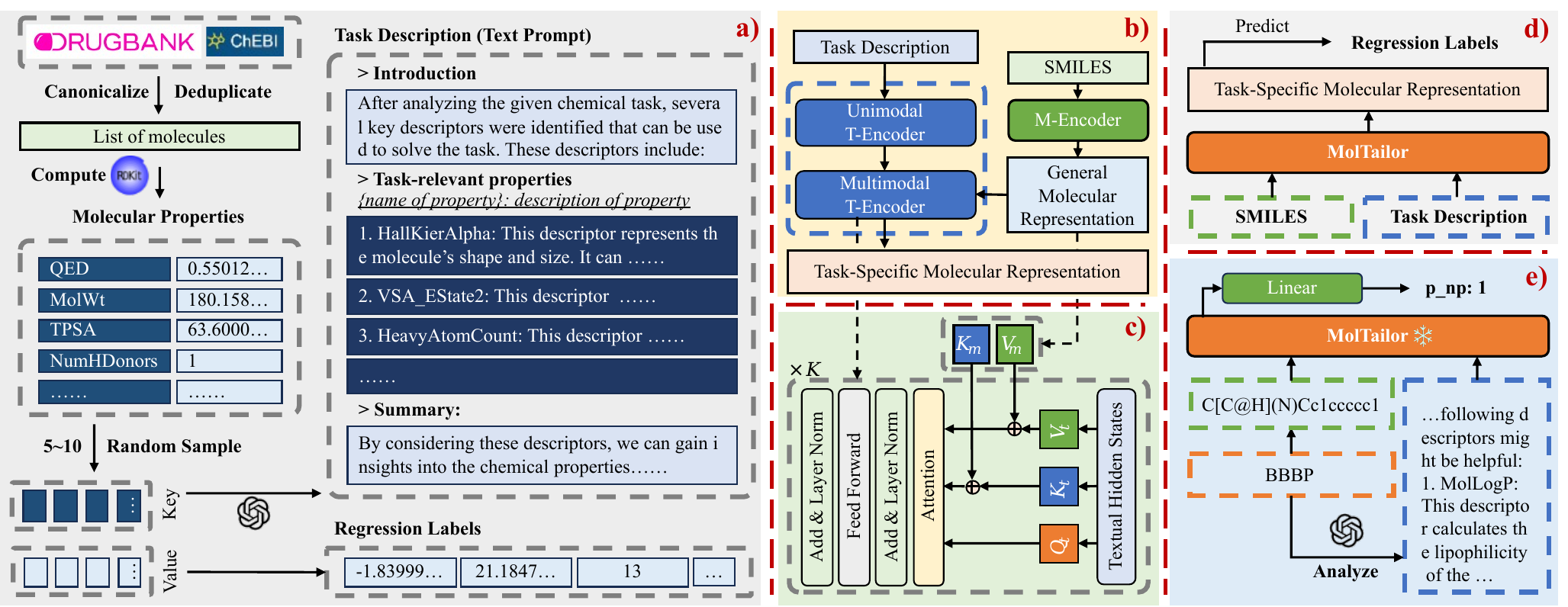} % Reduce the figure size so that it is slightly narrower than the column. Don't use precise values for figure width.This setup will avoid overfull boxes.
    \caption{Overview of the MolTailor framework. {a) The construction process of the MT-MTR dataset.} We obtain representative molecules from DrugBank \cite{wishart2018drugbank} and ChEBI \cite{hastings2016chebi}, and then use RDKit to calculate 209 properties for each molecule. For each molecule, we randomly sample 5-10 properties from the property set, use the property names to generate virtual task descriptions via GPT-3.5, and use the property values as regression labels. {b) Model architecture of MolTailor.} MolTailor consists of a language pretraining model (T-Encoder) and a molecular pretraining model (M-Encoder). The T-Encoder is divided into a unimodal part (for understanding task descriptions) and a multimodal part (for adjusting molecular representations). {c) Internal structure of the Multimodal T-Encoder.} It modifies the original Transformer Encoder Block to perform self-attention and cross-attention operations simultaneously: mapping the general molecular representation to obtain \(K_m\) and \(V_m\) vectors which are then concatenated with textual vectors \(K_t\) and \(V_t\). {d) Pretraining task of MolTailor.} The model needs to predict properties mentioned in the task description based on the molecule and text prompt. {e) Downstream tasks of MolTailor.} For a specific downstream task, we first generate the task description in the same format as pretraining via GPT-4 analysis, then take the SMILES and task description as input to predict labels for the corresponding task.}
    \label{fig:overview}
\end{figure*}

Fig. \ref{fig:overview} presents an overview of our proposed approach. In this section, we introduce MolTailor in three aspects: 1. Construction of the MT-MTR corpus. 2. Model architecture of MolTailor. 3. Pretraining of MolTailor and application of MolTailor to downstream tasks.

\subsection{Construction of MT-MTR}
The construction process of the MT-MTR (Molecule-Text Multi-Task Regression) dataset consists of 3 main steps as shown in Fig. \ref{fig:overview}a:

\textbf{Step 1: Obtain representative molecules.} We obtain molecules from DrugBank \cite{wishart2018drugbank} and ChEBI \cite{hastings2016chebi}, then use RDKit\footnote{https://www.rdkit.org} to deduplicate and canonicalize, resulting in 55,759 valid SMILES.

\textbf{Step 2: Calculate molecular properties.}  For each molecule obtained in the previous step, we use RDKit to calculate 209 properties\footnote{RDKit's ``Descriptor'' module provides 209 descriptors, which we uniformly refer to as ``properties'' in this paper for clarity.}. Among these properties, some are continuous values while others are discrete values. We unify them into regression tasks.

\textbf{Step 3: Obtain task descriptions and regression labels.} For each molecule, we randomly sample 5-10 properties from the property set obtained in the previous step. The sampled property names are filled into the prompt template in Tab. \ref{tab:prompt} (Prompt for Pretraining) to obtain text input fed into GPT for generating virtual task description. Since GPT-4 and GPT-3.5 achieve similar performance on this task, we use the more cost-effective GPT-3.5 for generation in our experiments. The values of the sampled properties are used as the regression labels. 

The resulting MT-MTR corpus contains (molecule, task description, regression labels) triplets. It is also worth noting that MT-MTR does not ask the model to predict all properties of the molecules; it only needs to predict properties mentioned in the text prompt. This aims to teach the model to generate molecule representations tailored to the text for improved predictions.

% Table generated by Excel2LaTeX from sheet 'Sheet3'
\begin{table}[htbp]
    \centering
    \begin{tabularx}{\columnwidth}{p{1cm}|X}
      \toprule
      \midrule
       & \textbf{Template} \\
      \midrule
      Prompt \newline{}for \newline{}Pre-\newline{}training & As a seasoned expert in the field of chemistry, your task is to analyse a chemical task. And you found following properties of chemical compounds can help solve this task. Please summarize your analysis. The length should be less than 300 tokens.\newline{}\textbf{Properties}: \newline{}\{\textit{Sampled Properties}\}. \newline{}\textbf{Task analysis results}: \\
      \midrule
      Prompt \newline{}for \newline{}Down-\newline{}stream & \textbf{Example}:\newline{}\{\textit{Example of task description from MT-MTR}\}\newline{}\textbf{Property Names}:\newline{}\{\textit{209 property names from RDKit}\}\newline{}Please analyze the {Task Name}. When discussing the properties related to this task, list any properties that you think may be helpful for solving the task. Simply provide the analysis results directly in less than 400 tokens, referring to the example for guidance. \\
      \midrule
      \bottomrule
    \end{tabularx}%
    \caption{Prompt templates for interacting with GPT to generate task descriptions.}
    \label{tab:prompt}%
  \end{table}%
  
\subsection{Model Architecture of MolTailor} 
As illustrated in Fig. \ref{fig:overview}b\&c, MolTailor consists of a language pretraining model (T-Encoder) and a molecular pretraining model (M-Encoder). The language model is divided into a Unimodal Text Encoder (UT-Encoder) and a Multimodal Text Encoder (MT-Encoder) by introducing a few parameters. We treat the M-Encoder as a knowledge base and the T-Encoder as an agent. The agent obtains the final desired representation by understanding natural language and adjusting the well-initialized molecular representation from the knowledge base. The UT-Encoder captures semantic information and the MT-Encoder then tailors molecular representations based on understanding task descriptions. 

Since any molecular pretraining model can serve as the M-Encoder in our method, in this section we mainly describe the Transformer encoder block (TEB), the fundamental component composing mainstream language models, as well as the internal structure of the MT-Encoder. Additionally, in our experiments, we tested two types of M-Encoder: CHEM-BERT \cite{kim2021merged} and ChemBERTa-2 \cite{ahmad2022chemberta}.

\subsubsection{Transformer Encoder Block.} As the basic building block of the Transformer \cite{vaswani2017attention} encoder, it consists of four main components: Multi-Head Self-Attention (MHA), Feed Forward Network (FFN), and Residual Connection with Layer Normalization (LN). For MHA, given hidden states \(\boldsymbol{x} \in \mathbb{R}^{(bs \times n \times d)}\) where \(bs\) indicates batch size, \(n\) indicates sequence length, and \(d\) indicates embedding dimension, it is first mapped to vectors \(\boldsymbol{Q}\), \(\boldsymbol{K}\) and \(\boldsymbol{V}\) by matrices \(\boldsymbol{W}_Q, \boldsymbol{W}_K, \boldsymbol{W}_V \in \mathbb{R}^{(d \times d)}\), keeping dimensions unchanged. Then, with \(h\) heads, \(d\) is split into \(h\) parts. After that, vectors are transposed and reshaped to get \(\boldsymbol{Q}, \boldsymbol{K}, \boldsymbol{V} \in \mathbb{R}^{(bs \times h \times n \times d/h)}\). Finally, attention is computed:
\begin{equation}
    \Attn(\boldsymbol{Q}, \boldsymbol{K}, \boldsymbol{V}) = \text{softmax}\left( \frac{\boldsymbol{Q}\boldsymbol{V}{^{\boldsymbol{T}}}}{\sqrt{d_k}} \right)
    \label{eq:1}
\end{equation}
\begin{equation}
    \MHA(\boldsymbol{x}) = \Attn(f_{Q}^r(\boldsymbol{x}), f_{K}^r(\boldsymbol{x}), f_{V}^r(\boldsymbol{x}))
    \label{eq:2}
\end{equation}
where \(f_*^r\) means \(f^{reshape}(\boldsymbol{x}\boldsymbol{W}_*)\). The output \(\boldsymbol{x}^* \in \mathbb{R}^{(bs \times n \times d)}\) of the MHA goes through the FFN, residual connection and LN to obtain the final output \(\boldsymbol{x} \in \mathbb{R}^{(bs \times n \times d)}\).
\begin{equation}
    \FFN(\boldsymbol{x}) = \mathrm{ReLU}(\boldsymbol{x}\boldsymbol{W}_1 + \boldsymbol{b}_1)\boldsymbol{W}_2 + \boldsymbol{b}_2
    \label{eq:3}
\end{equation}
\begin{equation}
    \boldsymbol{x} = \LN(\boldsymbol{x} + \FFN(\MHA(\boldsymbol{x})))
    \label{eq:4}
\end{equation}

\subsubsection{Multimodal T-Encoder.} Since the task descriptions contain biomedical terminology, we chose PubMedBERT \cite{gu2021domain} as the backbone for the T-Encoder. Additionally, we also present in the appendix the results of using BioLinkBERT \cite{yasunaga2022linkbert}. PubMedBERT contains 12 TEBs. We designate the first 9 layers as the UT-Encoder, and the remaining 3 layers as the MT-Encoder. 

To build the MT-Encoder, we draw on previous work \cite{chen2022hybrid} to modify the original $\MHA$ to $\MHA^*$ that can simultaneously perform Self-Attention and Cross-Attention operations, by introducing very few parameters. Specifically, the MT-Encoder takes $\boldsymbol{x}_{t}$ and $\boldsymbol{x}_{m}$ as inputs, where $\boldsymbol{x}_{t} \in \mathbb{R}^{(bs \times n_{t} \times d_{t})}$ denotes the hidden states from the UT-Encoder, and $\boldsymbol{x}_{m} \in \mathbb{R}^{(bs \times n_{m} \times d_{m})}$ denotes the general molecular representation (i.e., the last hidden states from the  M-Encoder). $\MHA^*$ computes as follows:
\begin{equation}
    \begin{aligned}
        \boldsymbol{Q}^*      & = f^r(\boldsymbol{x}_{t}\boldsymbol{W}_Q^t)                              \\
        \boldsymbol{K}^*      & = f^r([\boldsymbol{x}_{t}\boldsymbol{W}_K^t, \boldsymbol{x}_{m}\boldsymbol{W}_K^m])    \\
        \boldsymbol{V}^*      & = f^r([\boldsymbol{x}_{t}\boldsymbol{W}_V^t, \boldsymbol{x}_{m}\boldsymbol{W}_V^m])    \\
        \MHA^*(\boldsymbol{x}_t, \boldsymbol{x}_m) & = \Attn(\boldsymbol{Q}^*, \boldsymbol{K}^*, \boldsymbol{V}^*)
        \label{eq:5}
    \end{aligned}
\end{equation}
where $f^r$ denotes reshape function, $[*]$ denotes concatenation along the dimension corresponding to the sequence length $n$. The output of the MT-Encoder Block is computed:
\begin{equation}
    \boldsymbol{x}^* = \LN(\boldsymbol{x}_t + \FFN(\MHA^*(\boldsymbol{x}_t, \boldsymbol{x}_m))) 
    \label{eq:6}
\end{equation}
\citet{chen2022hybrid} formally proves that \(\MHA^*(\boldsymbol{x})\) is equivalent to a weighted average of self-attention and cross-attention:
\begin{equation}
    \begin{aligned}
    \MHA^*(\boldsymbol{x}_t, \boldsymbol{x}_m) & = (1-\lambda(\boldsymbol{x}_{t}))\MHA(\boldsymbol{x}_{t}) \\
                           & + \lambda(\boldsymbol{x}_{t})\MHA(\boldsymbol{Q}_{t}, \boldsymbol{K}_m, \boldsymbol{V}_m)
    \label{eq:7}
    \end{aligned}
\end{equation}

\subsection{Pretraining and Downstream Application}
\subsubsection{Pretraining on MT-MTR.} We use MT-MTR as the pretraining objective . In detail, given SMILES (\(\boldsymbol{z}_{m}\)) and task description (\(\boldsymbol{z}_{t}\)) as input, the model predicts regression labels \(\boldsymbol{y} \in \mathbb{R}^{(1 \times 209)}\), as displayed in Fig. \ref{fig:overview}d. The pretraining loss uses MSE and can be formulated as:
\begin{equation}
    \mathcal{L} = \frac{1}{N}\sum_{j=1}^{N}\left( \frac{1}{\mathrm{count}(\boldsymbol{m})}\sum_{i=1}^{M}m_{ij}(y_{ij}-\hat{y}_{ij})^2 \right)
\end{equation}
where \(N\) is the number of samples, \(\boldsymbol{m} \in \mathbb{R}^{(1 \times 209)}\) is a 0-1 vector, in which the value 1 indicates the existence of the corresponding regression label, while 0 indicates its absence, \(\mathrm{count}(\boldsymbol{m})\) means the number of valid labels, \(M\) is the number of all properties (here \(M = 209\)), \(y_{ij}\) is the ground truth regression label, and \(\hat{y}_{ij}\) is the predicted value.

\subsubsection{Downstream Application.}
When applying MolTailor to downstream tasks, we use GPT-4 to analyze the specific task and then generate the corresponding task description, as shown in Fig. \ref{fig:overview}e. We use the template in Tab. \ref{tab:prompt} to prompt GPT to generate descriptions that mimic the format of those in the pretraining corpus. Notably, we restrict it such that it can only select the names of properties from the set supported by RDKit. Then, we feed the generated analysis and SMILES into MolTailor to obtain the task-specific molecular representation:
\begin{equation}
    \boldsymbol{z} = \mathrm{MolTailor}\left(\boldsymbol{z}_{m} | \boldsymbol{z}_{t} \right)
\end{equation}
where $\boldsymbol{z}$ denotes the task-specific molecular representation. Finally, $\boldsymbol{z}$ goes through a prediction head to predict the labels corresponding to the task.

\section{Experiments}
In this section, we conduct comprehensive experiments to demonstrate the efficacy of our proposed method. The experiments are designed to analyze the method by addressing the following six key questions:

\textbf{Q1}: Are the task-specific molecular representations generated by MolTailor better than general representations?
\textbf{Q2}: Can MolTailor achieve performance improvements on different M-Encoders?
\textbf{Q3}: How do the task descriptions influence the effectiveness of pretraining?
\textbf{Q4}: How do different text prompts affect MolTailor?
\textbf{Q5}: When MolTailor achieves better performance, are the task-relevant properties in the molecular representations enhanced?
\textbf{Q6}: Does MolTailor pay attention to the key information in both the molecules and text prompts?

\subsection{Experimental Setup}
\subsubsection{Pretraining Corpus.} We use MT-MTR corpus for pretraining, which contains 55,759 triples. Moreover, we present the data overlap analysis results in the appendix.

\subsubsection{Downstream Datasets.} We select 8 representative tasks from MoleculeNet \cite{wu2018moleculenet} for experiments, which consist of 4 classification tasks (BBBP, ClinTox, HIV, Tox21) and 4 regression tasks (ESOL, FreeSolv, Lipophilicity, QM8), covering physiology, biophysics, physical chemistry, and quantum mechanics. Following \citet{wu2018moleculenet}, each task uses the recommended splitting method to divide data into training/validation/test sets with a ratio of 8:1:1.

\subsubsection{Baselines.} We adopt the following four types of baselines:
\begin{itemize}
    \item \textbf{Molecular Fingerprints}: MACCSFP \cite{durant2002reoptimization} encodes molecules based on substructures, RDKitFP \cite{o2016comparing} encodes molecules based on topology or path, and MorganFP \cite{rogers2010extended} encodes molecular environment and structure starting from atoms within a radius.
    \item \textbf{Sequence-based Methods}: BioLinkBERT\cite{yasunaga2022linkbert}, PubMedBERT \cite{gu2021domain}; ChemBERTa-2 \cite{ahmad2022chemberta}, and CHEM-BERT \cite{kim2021merged}.
    \item \textbf{Graph-based Methods}: Grover \cite{rong2020self}, MolCLR \cite{wang2022molecular}, Mole-BERT \cite{xia2022mole}, and Uni-Mol \cite{zhou2023uni}.
    \item \textbf{Multimodal Methods}: KCL \cite{fang2022molecular}, KV-PLM \cite{zeng2022deep}, MolT5 \cite{edwards2022translation}, and MoMu \cite{su2022molecular}.
\end{itemize}
We additionally construct Random and RDKit-DP as baselines for comparison, where Random refers to random vectors, and RDKit-DP consists of the 209 molecular properties calculated by RDKit.

\subsubsection{Evaluation Methodology.}
To better evaluate the molecular representations learned by different models, we conduct experiments in linear probe setting. As a result, the baseline experimental results reported in this paper may differ from the original papers. We freeze the model parameters to extract representations for downstream tasks. These extracted representations are then mapped to labels through a learnable linear layer. 

Following the recommendation of \citet{wu2018moleculenet}, we use ROC-AUC as the evaluation metric for classification tasks. For the regression task qm8, we use MAE, and for other regression tasks, we use RMSE. To ensure fairness, we use Optuna \cite{akiba2019optuna} to search 10 learning rates (LRs) for each model. We repeat each task 3 times and report the mean and standard deviation. Due to space limitations, the standard deviations are included in the appendix.

\subsubsection{Implementation Details.}
For the pretraining phase, we employ the AdamW optimizer, complemented by a linear learning rate scheduler. We set the LR at 5.5e-5 and use a warmup ratio of 0.1. The training is conducted 50 epochs with a batch size of 64, utilizing two A100-SXM4-80GB GPUs. For the downstream tasks, we opt for the Adam optimizer and leverage Optuna for hyperparameter tuning, conducting 10 trials to identify the optimal LR within the range of 1e-5 to 1e-2 for each model on every task. The optimal LR is determined based on the performance of the validation set. We train our model using a batch size of 64 on a single GeForce RTX 2080 Ti GPU, employing an early stop mechanism with a patience setting of 3 and limiting the training to a maximum of 50 epochs.

\renewcommand{\arraystretch}{0.8}
\begin{table*}[t]
    \centering
      \begin{tabular}{l|cccc|cccc|r}
      \toprule
      \midrule
      \textbf{Models} & \multicolumn{4}{c|}{\textbf{Classification (ROC-AUC)}} & \multicolumn{4}{c|}{\textbf{Regression (RMSE / MAE)}} & \multirow{5}[4]{*}{\textbf{Params}} \\
  \cmidrule{1-9}    Dataset & BBBP  & ClinTox & HIV   & Tox21 & ESOL  & FreeSolv & Lip   & QM8   &  \\
      \#Molecules & 2039  & 1478  & 41127 & 7831  & 1128  & 642   & 4200  & 21786 &  \\
      \#Split & Scaffold & Random & Scaffold & Random & Random & Random & Random & Random &  \\
      \#Tasks & 1     & 2     & 1     & 12    & 1     & 1     & 2     & 16    &  \\
      \midrule
      Random & 48.38  & 56.01  & 49.54  & 51.11  & 3.3358  & 5.4831  & 1.3813  & 0.0320  & - \\
      RDKit-DP & 78.25  & 67.36  & 70.85  & 65.61  & 4.8940  & 2.8068  & 0.9963  & 0.0202  & - \\
      \midrule
      RDKit-FP & 87.65  & 57.13  & 78.66  & 76.14  & 1.0830  & 2.0725  & 0.9007  & \textbf{0.0181} & - \\
      MACCS-FP & 81.64  & 83.05  & 77.53  & 77.27  & 1.0833  & 1.9086  & 0.9886  & 0.0196  & - \\
      Morgan-FP & 82.73  & 72.61  & \textbf{82.65} & 75.29  & 1.2413  & 2.1896  & 0.8196  & 0.0200  & - \\
      \midrule
      Grover & 79.83  & 87.75  & 77.47  & 79.61  & 0.8977  & 1.9041  & 0.8301  & 0.0184  & 107M \\
      MolCLR & 81.27  & 78.15  & 71.48  & 75.61  & 1.3421  & 3.0436  & 1.0448  & 0.0219  & 2M \\
      Mole-BERT & 82.70  & 81.82  & 79.35  & \textbf{84.20} & 1.1379  & 2.3626  & 0.8316  & 0.0221  & 2M \\
      Uni-Mol & 79.52  & 88.65  & 74.18  & 78.08  & 1.0509  & 2.6913  & 1.0363  & 0.0219  & 48M \\
      \midrule
      BioLinkBERT & 83.81  & 87.75  & 71.24  & 73.81  & 1.1739  & 3.1350  & 1.0589  & 0.0234  & 108M \\
      PubMedBERT & \textbf{89.10} & 84.29  & 72.30  & 73.77  & 1.0715  & 2.5999  & 1.0851  & 0.0232  & 108M \\
      ChemBERTa-2 & 84.70  & 84.21  & 78.88  & 80.75  & 0.8103  & 1.8439  & 0.7948  & 0.0191  & 3M \\
      CHEM-BERT & 84.10  & \textbf{93.80} & 76.99  & 80.54  & 0.7973  & 2.0214  & 0.8571  & 0.0215  & 51M \\
      \midrule
      KCL   & 76.86  & 60.80  & 68.48  & 74.98  & 0.8728  & 2.7615  & 0.9868  & 0.0225  & 1M \\
      KV-PLM & 86.36  & 81.20  & 73.52  & 74.62  & 1.1785  & 2.8840  & 1.1004  & 0.0233  & 109M \\
      MoMu-ME & 80.41  & 67.99  & 71.91  & 74.75  & 1.4135  & 2.3229  & 0.9835  & 0.0222  & 2M \\
      MoMu-TE & 82.24  & 81.94  & 67.88  & 73.07  & 1.2562  & 3.1480  & 1.0885  & 0.0250  & 109M \\
      \midrule
      MolTailor$^{*}$ & 84.65  & \underline{85.95}  & 76.42  & 80.32  & \underline{\textbf{0.7128}} & \underline{\textbf{1.7826}} & \underline{\textbf{0.7848}} & \underline{0.0185}  & 112M \\
      MolTailor & 81.15  & 92.37  & \underline{77.42}  & \underline{80.67}  & \underline{0.7234}  & \underline{1.7881}  & \underline{0.8107}  & \underline{0.0196}  & 161M \\
      \midrule
      \bottomrule
      \end{tabular}%
    \caption{Evaluation results of MolTailor and baselines under the linear probe setting on MoleculeNet. Here, ME stands for Molecule Encoder and TE stands for Text Encoder in MoMu-ME/TE. MolTailor$^{*}$ denotes using ChemBERTa-2 as the M-Encoder, while MolTailor indicates using CHEM-BERT. The table shows the average performance over 3 runs. Standard deviations are omitted due to space limitations but can be viewed in the appendix. Results in bold indicate state-of-the-art (SOTA) performance, while the underlined results show that the model outperforms the backbone used in the M-Encoder.}
    \label{tab:main-result}%
\end{table*}%
\renewcommand{\arraystretch}{1}

\subsection{Performance Comparison (Q1 \& Q2)}
\textbf{Q1:} We evaluate whether molecular representations enhanced by task descriptions could improve performance across 8 tasks. Tab. \ref{tab:main-result} shows that MolTailor achieves performance gains over the backbone model on the 4 regression tasks, and notably attains state-of-the-art (SOTA) results on ESOL, FreeSolv, and Lipophilicity datasets. On the 4 classification tasks, MolTailor's performance is inconsistent, with gains on HIV and Tox21 but losses on the other 2 datasets when using CHEM-BERT as backbone. We hypothesize that the difference in model performance on classification and regression tasks is due to the pretraining data benefiting the regression task more. Meanwhile, we experimented with converting the regression pretraining task into a classification form. The results were still similar, eliminating the influence of the form of the pretraining task on downstream tasks.

\textbf{Q2:} We experiment on two backbones, ChemBERTa-2 and CHEM-BERT. MolTailor achieves similar gains on top of both backbones. Notably, the performance differences between the backbones are also mirrored in the respective MolTailor variants. This demonstrates MolTailor's transferability and ability to inherit strengths of different backbones. The results support that further gains may be achievable by transferring MolTailor to new state-of-the-art single-modality models.

\subsection{Ablation Study (Q3 \& Q4 \& Q5)}
\textbf{Q3. } To evaluate the influence of task descriptions on pretraining in MT-MTR, we first remove the task descriptions to obtain the dataset MT-MTR$^*$, then pretrain CHEM-BERT and PubMedBERT on MT-MTR$^*$ to get CHEM-BERT$^*$ and PubMedBERT$^*$ models. Next, we concatenate the task descriptions after the SMILES strings to construct dataset MT-MTR$^\dagger$, and pretrain PubMedBERT on MT-MTR$^\dagger$ to obtain PubMedBERT$^\dagger$. As shown in Tab. \ref{tab:q3}, CHEM-BERT$^*$ improves over untrained CHEM-BERT on regression tasks but drops on classification, indicating MTR as a pretraining task benefits downstream regression but negatively impacts classification. Also, PubMedBERT$^\dagger$ over PubMedBERT$^*$ and MolTailor over CHEM-BERT both demonstrate further regression performance gains but classification performance drops. This shows introducing task descriptions helps models better learn from the data. That is, if constructing new labels can improve model performance on classification tasks, then supplementing task descriptions can further amplify such gains.
% Table generated by Excel2LaTeX from sheet 'Sheet5'

\textbf{Q4. } We replace the task-specific descriptions generated via GPT-4 with the irrelevant text \textit{``to be or not to be, this is the question.''} as noise prompts. Results in Tab. \ref{tab:q4} show degraded performance with noise prompts, indicating information in the task descriptions does help the model generate better molecular representations. However, it should be noted that the performance drop is slight, suggesting textual information is not as important as expected in the process of generating representations. Further investigation into this phenomenon is needed.
% Table generated by Excel2LaTeX from sheet 'Sheet6'

\renewcommand{\arraystretch}{0.8}
\begin{table}[htbp]
    \centering
      \begin{tabular}{lll}
      \toprule
      \midrule
      \textbf{Models} & \textbf{Classification$\uparrow$} & \textbf{Regression$\downarrow$} \\
      \midrule
      CHEM-BERT & 83.86 & 0.9243 \\
      CHEM-BERT$^*$ & 83.11 & \underline{0.9209} \\
      \midrule
      PubMedBERT & 79.87 & 1.1949 \\
      PubMedBERT$^*$ & \underline{81.67}& \underline{0.9148} \\
      PubMedBERT$^\dagger$ & 81.08 & \underline{0.9131} \\
      \midrule
      MolTailor & 82.09 & \underline{0.8475}\\
      \midrule
      \bottomrule
      \end{tabular}%
    \caption{Experimental results for Q3. CHEM-BERT and PubMedBERT use their original weights when evaluated on downstream tasks. The remaining methods are pretrained on MT-MTR and its variants, with overlapping molecules between pretraining and downstream data removed. Additionally, all methods are pretrained for only 20 epochs.}
    \label{tab:q3}%
\end{table}%
\renewcommand{\arraystretch}{1}

\renewcommand{\arraystretch}{0.8}
\begin{table}[htbp]
    \centering
      \begin{tabular}{lll}
      \toprule
      \toprule
      \textbf{Prompt Types} & \textbf{Classification$\uparrow$} & \textbf{Regression$\downarrow$} \\
      \midrule
      Origin & 82.92  & 0.8348 \\
      Noise & 82.77  & 0.8541 \\
      \bottomrule
      \bottomrule
      \end{tabular}%
    \caption{Experimental results for Q4.}
    \label{tab:q4}%
\end{table}%
\renewcommand{\arraystretch}{1}

\begin{figure}[t]
    \centering
    \includegraphics[width=1\columnwidth]{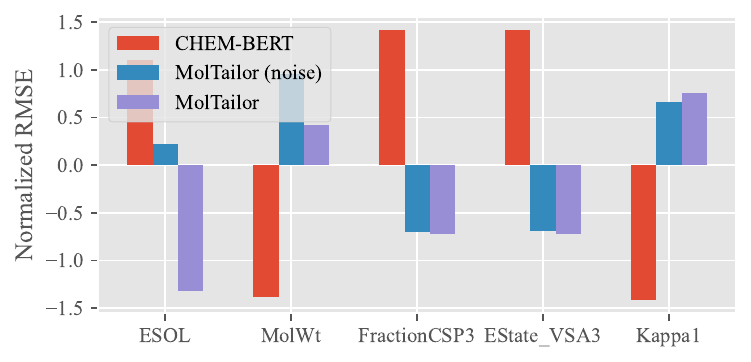} % Reduce the figure size so that it is slightly narrower than the column. Don't use precise values for figure width.This setup will avoid overfull boxes.
    \caption{Performance of three methods on ESOL and molecular properties prediction tasks for Q5. The x-axis shows the task names, the y-axis shows the normalized RMSE, with lower values indicating better performance. Of the four molecular properties, the first three are related to the ESOL task, while the last one is opposite.}
    \label{fig:q5}
\end{figure}

\textbf{Q5. }The experiments in Tab. \ref{tab:main-result} show that the molecule representations generated by MolTailor attained better performance. However, are the obtained representations more task-specific, meaning the task-relevant properties in the representations are enhanced? To validate this, we designed the following experiment: we used the representations generated by different methods to predict task-related and task-unrelated properties. If the expectations are met, then the representations enhanced by MolTailor should have better performance in predicting task-related molecular properties.

We conduct experiments on the ESOL dataset. For the models, we use MolTailor and its corresponding backbone CHEM-BERT. We also test the performance of MolTailor using irrelevant text as the prompt, denoted as MolTailor(noise). In detail, we first use RDKit to compute the properties of the molecules in the ESOL dataset, obtaining the ESOL-MTR dataset. Then, we randomly split the dataset into 8:1:1 for training, validation, and testing. 

The results are shown in Fig. \ref{fig:q5}. We present the performance of the three methods on the ESOL task and in predicting four molecular properties. Among the four selected properties, two have names mentioned in the prompt and are relevant to the ESOL task (MolWt and FractionCSP3), one is not mentioned but is relevant (EState\_VSA3), and one is neither mentioned nor relevant (Kappa1). We use RMSE as the evaluation metric. For better visualization, we normalize the RMSE under each task using the formula: $\text{Normalized RMSE} = \tfrac{\text{RMSE} - \text{mean}(\text{RMSE})}{\text{std}(\text{RMSE})}$.

The results show that: a) Compared to CHEM-BERT, MolTailor does focus more on the properties mentioned in the prompt, validating that the obtained representations are indeed task-specific. b) MolTailor not only pays more attention to the relevant properties explicitly stated in the prompt, but also to the unstated yet still relevant properties. Meanwhile, it decreases attention on the unrelated, unmentioned properties. This demonstrates the method's generalization capability. c) The declined performance of MolTailor(noise) in property prediction indicates that the prompt does help the model attend to critical information.

\subsection{Case Study (Q6)}
\textbf{Q6.} We analyze the attention matrices from the last layers of the UT/MT-Encoder to investigate whether MolTailor pays attention to key information. We select the ESOL dataset, which is related to solubility, to conduct experiments. If the model attends to information such as molecular weight or polar functional groups that are critical determinants of solubility, it suggests key information is captured.

In detail, We pass the SMILES from the test set through MolTailor trained on ESOL to obtain the required attention matrices $\boldsymbol{M}_{UT}$ and $\boldsymbol{M}_{MT}$. Then, we take the values of the matrix between the ``[CLS]'' token and the other tokens as the required attention weights.

\begin{figure}[t]
    \centering
    \includegraphics[width=1\columnwidth]{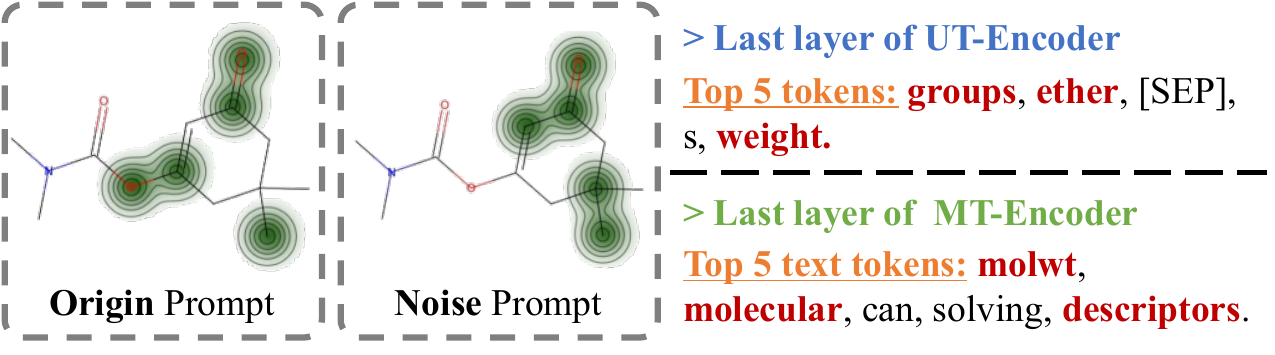} % Reduce the figure size so that it is slightly narrower than the column. Don't use precise values for figure width.This setup will avoid overfull boxes.
    \caption{Visualization of the attention, answering {Q6}. The two molecular graphs on the left show MolTailor's attention over the molecules under different prompts. The text on the right shows which input tokens from the original prompt MolTailor pays most attention to.}
    \label{fig:attention}
\end{figure}

The analysis results in Fig. \ref{fig:attention} show that MolTailor does pay attention to solubility-related information, such as ``groups'', ``ethers'', and other highlighted tokens. Notably, the tokens here are merged to form complete words. Additionally, the model also pays attention, as seen in the left molecule attention graph, to polar functional groups within the SMILES. Furthermore, variations between the two molecule attention graphs under different prompts indicate effective prompts help the model better focus on key information that assist in solving the task

\section{Conclusion and Future Work}
Overall, in this work we propose a new perspective on molecular representation learning. Instead of trying to incorporate more information into the representations, we obtain more task-specific representations by combining contextual information. At the same time, we not only utilize the knowledge contained in the text modality, but also try to leverage the reasoning ability of language models, which has huge potential in the era of large language models.

In the future, we first plan to explore new pretraining tasks that can stably improve model performance on both classification and regression tasks. Secondly, we will explore how to build molecular-text multimodal models based on large language models. Finally, we look forward to in-depth collaborations with domain experts to apply the molecular-text multimodal methods to actual production problems.

\section{Acknowledgements}
We thank the reviewers for their detailed and constructive suggestions and for the affirmation of our paper. Meanwhile, we gratefully acknowledge the support of the National Key R\&D Program of China (2021ZD0113302), the National Natural Science Foundation of China Youth Fund (62206079), and the Heilongjiang Provincial Natural Science Foundation of China (YQ2022F006). We also appreciate the support from Du Xiaoman (Beijing) Science Technology Co., Ltd. @ on our research.

\bibliography{aaai24}

\begin{thebibliography}{53}
\providecommand{\natexlab}[1]{#1}

\bibitem[{Ahmad et~al.(2022)Ahmad, Simon, Chithrananda, Grand, and Ramsundar}]{ahmad2022chemberta}
Ahmad, W.; Simon, E.; Chithrananda, S.; Grand, G.; and Ramsundar, B. 2022.
\newblock Chemberta-2: Towards chemical foundation models.
\newblock \emph{arXiv preprint arXiv:2209.01712}.

\bibitem[{Akiba et~al.(2019)Akiba, Sano, Yanase, Ohta, and Koyama}]{akiba2019optuna}
Akiba, T.; Sano, S.; Yanase, T.; Ohta, T.; and Koyama, M. 2019.
\newblock Optuna: A next-generation hyperparameter optimization framework.
\newblock In \emph{Proceedings of the 25th ACM SIGKDD international conference on knowledge discovery \& data mining}, 2623--2631.

\bibitem[{Bran et~al.(2023)Bran, Cox, Schilter, Baldassari, White, and Schwaller}]{bran2023augmenting}
Bran, A.~M.; Cox, S.; Schilter, O.; Baldassari, C.; White, A.; and Schwaller, P. 2023.
\newblock Augmenting large language models with chemistry tools.
\newblock In \emph{NeurIPS 2023 AI for Science Workshop}.

\bibitem[{Chen et~al.(2022)Chen, Zhang, Li, Deng, Tan, Xu, Huang, Si, and Chen}]{chen2022hybrid}
Chen, X.; Zhang, N.; Li, L.; Deng, S.; Tan, C.; Xu, C.; Huang, F.; Si, L.; and Chen, H. 2022.
\newblock Hybrid transformer with multi-level fusion for multimodal knowledge graph completion.
\newblock In \emph{Proceedings of the 45th International ACM SIGIR Conference on Research and Development in Information Retrieval}, 904--915.

\bibitem[{Chen et~al.(2023)Chen, Xi, Du, Wang, Jianyu, Zhao, and Qin}]{chen2023artificially}
Chen, Y.; Xi, N.; Du, Y.; Wang, H.; Jianyu, C.; Zhao, S.; and Qin, B. 2023.
\newblock From Artificially Real to Real: Leveraging Pseudo Data from Large Language Models for Low-Resource Molecule Discovery.
\newblock \emph{arXiv preprint arXiv:2309.05203}.

\bibitem[{Chithrananda, Grand, and Ramsundar(2020)}]{chithrananda2020chemberta}
Chithrananda, S.; Grand, G.; and Ramsundar, B. 2020.
\newblock ChemBERTa: large-scale self-supervised pretraining for molecular property prediction.
\newblock \emph{arXiv preprint arXiv:2010.09885}.

\bibitem[{Christofidellis et~al.(2023)Christofidellis, Giannone, Born, Winther, Laino, and Manica}]{christofidellis2023unifying}
Christofidellis, D.; Giannone, G.; Born, J.; Winther, O.; Laino, T.; and Manica, M. 2023.
\newblock Unifying molecular and textual representations via multi-task language modelling.
\newblock \emph{arXiv preprint arXiv:2301.12586}.

\bibitem[{Devlin et~al.(2018)Devlin, Chang, Lee, and Toutanova}]{devlin2018bert}
Devlin, J.; Chang, M.-W.; Lee, K.; and Toutanova, K. 2018.
\newblock Bert: Pre-training of deep bidirectional transformers for language understanding.
\newblock \emph{arXiv preprint arXiv:1810.04805}.

\bibitem[{Durant et~al.(2002)Durant, Leland, Henry, and Nourse}]{durant2002reoptimization}
Durant, J.~L.; Leland, B.~A.; Henry, D.~R.; and Nourse, J.~G. 2002.
\newblock Reoptimization of MDL keys for use in drug discovery.
\newblock \emph{Journal of chemical information and computer sciences}, 42(6): 1273--1280.

\bibitem[{Edwards et~al.(2022)Edwards, Lai, Ros, Honke, Cho, and Ji}]{edwards2022translation}
Edwards, C.; Lai, T.; Ros, K.; Honke, G.; Cho, K.; and Ji, H. 2022.
\newblock Translation between molecules and natural language.
\newblock \emph{arXiv preprint arXiv:2204.11817}.

\bibitem[{Fang et~al.(2022)Fang, Zhang, Yang, Zhuang, Deng, Zhang, Qin, Chen, Fan, and Chen}]{fang2022molecular}
Fang, Y.; Zhang, Q.; Yang, H.; Zhuang, X.; Deng, S.; Zhang, W.; Qin, M.; Chen, Z.; Fan, X.; and Chen, H. 2022.
\newblock Molecular contrastive learning with chemical element knowledge graph.
\newblock In \emph{Proceedings of the AAAI Conference on Artificial Intelligence}, volume~36, 3968--3976.

\bibitem[{Gilmer et~al.(2017)Gilmer, Schoenholz, Riley, Vinyals, and Dahl}]{gilmer2017neural}
Gilmer, J.; Schoenholz, S.~S.; Riley, P.~F.; Vinyals, O.; and Dahl, G.~E. 2017.
\newblock Neural message passing for quantum chemistry.
\newblock In \emph{International conference on machine learning}, 1263--1272. PMLR.

\bibitem[{Gu et~al.(2021)Gu, Tinn, Cheng, Lucas, Usuyama, Liu, Naumann, Gao, and Poon}]{gu2021domain}
Gu, Y.; Tinn, R.; Cheng, H.; Lucas, M.; Usuyama, N.; Liu, X.; Naumann, T.; Gao, J.; and Poon, H. 2021.
\newblock Domain-specific language model pretraining for biomedical natural language processing.
\newblock \emph{ACM Transactions on Computing for Healthcare (HEALTH)}, 3(1): 1--23.

\bibitem[{Guo et~al.(2021)Guo, Sharma, Martinez, Du, and Abraham}]{guo2021multilingual}
Guo, Z.; Sharma, P.; Martinez, A.; Du, L.; and Abraham, R. 2021.
\newblock Multilingual molecular representation learning via contrastive pre-training.
\newblock \emph{arXiv preprint arXiv:2109.08830}.

\bibitem[{Hastings et~al.(2016)Hastings, Owen, Dekker, Ennis, Kale, Muthukrishnan, Turner, Swainston, Mendes, and Steinbeck}]{hastings2016chebi}
Hastings, J.; Owen, G.; Dekker, A.; Ennis, M.; Kale, N.; Muthukrishnan, V.; Turner, S.; Swainston, N.; Mendes, P.; and Steinbeck, C. 2016.
\newblock ChEBI in 2016: Improved services and an expanding collection of metabolites.
\newblock \emph{Nucleic acids research}, 44(D1): D1214--D1219.

\bibitem[{Hu et~al.(2019)Hu, Liu, Gomes, Zitnik, Liang, Pande, and Leskovec}]{hu2019strategies}
Hu, W.; Liu, B.; Gomes, J.; Zitnik, M.; Liang, P.; Pande, V.; and Leskovec, J. 2019.
\newblock Strategies for pre-training graph neural networks.
\newblock \emph{arXiv preprint arXiv:1905.12265}.

\bibitem[{Kim et~al.(2021)Kim, Lee, Ahn, and Lee}]{kim2021merged}
Kim, H.; Lee, J.; Ahn, S.; and Lee, J.~R. 2021.
\newblock A merged molecular representation learning for molecular properties prediction with a web-based service.
\newblock \emph{Scientific Reports}, 11(1): 11028.

\bibitem[{Kim, Son, and Kim(2021)}]{kim2021vilt}
Kim, W.; Son, B.; and Kim, I. 2021.
\newblock Vilt: Vision-and-language transformer without convolution or region supervision.
\newblock In \emph{International Conference on Machine Learning}, 5583--5594. PMLR.

\bibitem[{Li et~al.(2019)Li, Yatskar, Yin, Hsieh, and Chang}]{li2019visualbert}
Li, L.~H.; Yatskar, M.; Yin, D.; Hsieh, C.-J.; and Chang, K.-W. 2019.
\newblock Visualbert: A simple and performant baseline for vision and language.
\newblock \emph{arXiv preprint arXiv:1908.03557}.

\bibitem[{Liu, Guo, and Tang(2022)}]{liu2022molecular}
Liu, S.; Guo, H.; and Tang, J. 2022.
\newblock Molecular geometry pretraining with se (3)-invariant denoising distance matching.
\newblock \emph{arXiv preprint arXiv:2206.13602}.

\bibitem[{Liu et~al.(2022)Liu, Nie, Wang, Lu, Qiao, Liu, Tang, Xiao, and Anandkumar}]{liu2022multi}
Liu, S.; Nie, W.; Wang, C.; Lu, J.; Qiao, Z.; Liu, L.; Tang, J.; Xiao, C.; and Anandkumar, A. 2022.
\newblock Multi-modal molecule structure-text model for text-based retrieval and editing.
\newblock \emph{arXiv preprint arXiv:2212.10789}.

\bibitem[{Liu et~al.(2021)Liu, Wang, Liu, Lasenby, Guo, and Tang}]{liu2021pre}
Liu, S.; Wang, H.; Liu, W.; Lasenby, J.; Guo, H.; and Tang, J. 2021.
\newblock Pre-training molecular graph representation with 3d geometry.
\newblock \emph{arXiv preprint arXiv:2110.07728}.

\bibitem[{Liu et~al.(2023)Liu, Zhang, Xia, Wu, Xie, Qin, Zhang, and Liu}]{liu2023molxpt}
Liu, Z.; Zhang, W.; Xia, Y.; Wu, L.; Xie, S.; Qin, T.; Zhang, M.; and Liu, T.-Y. 2023.
\newblock MolXPT: Wrapping Molecules with Text for Generative Pre-training.
\newblock \emph{arXiv preprint arXiv:2305.10688}.

\bibitem[{Lu et~al.(2019)Lu, Batra, Parikh, and Lee}]{lu2019vilbert}
Lu, J.; Batra, D.; Parikh, D.; and Lee, S. 2019.
\newblock Vilbert: Pretraining task-agnostic visiolinguistic representations for vision-and-language tasks.
\newblock \emph{Advances in neural information processing systems}, 32.

\bibitem[{Mamoshina et~al.(2016)Mamoshina, Vieira, Putin, and Zhavoronkov}]{mamoshina2016applications}
Mamoshina, P.; Vieira, A.; Putin, E.; and Zhavoronkov, A. 2016.
\newblock Applications of deep learning in biomedicine.
\newblock \emph{Molecular pharmaceutics}, 13(5): 1445--1454.

\bibitem[{O’Boyle and Sayle(2016)}]{o2016comparing}
O’Boyle, N.~M.; and Sayle, R.~A. 2016.
\newblock Comparing structural fingerprints using a literature-based similarity benchmark.
\newblock \emph{Journal of cheminformatics}, 8(1): 1--14.

\bibitem[{Radford et~al.(2021)Radford, Kim, Hallacy, Ramesh, Goh, Agarwal, Sastry, Askell, Mishkin, Clark et~al.}]{radford2021learning}
Radford, A.; Kim, J.~W.; Hallacy, C.; Ramesh, A.; Goh, G.; Agarwal, S.; Sastry, G.; Askell, A.; Mishkin, P.; Clark, J.; et~al. 2021.
\newblock Learning transferable visual models from natural language supervision.
\newblock In \emph{International conference on machine learning}, 8748--8763. PMLR.

\bibitem[{Rogers and Hahn(2010)}]{rogers2010extended}
Rogers, D.; and Hahn, M. 2010.
\newblock Extended-connectivity fingerprints.
\newblock \emph{Journal of chemical information and modeling}, 50(5): 742--754.

\bibitem[{Rong et~al.(2020)Rong, Bian, Xu, Xie, Wei, Huang, and Huang}]{rong2020self}
Rong, Y.; Bian, Y.; Xu, T.; Xie, W.; Wei, Y.; Huang, W.; and Huang, J. 2020.
\newblock Self-supervised graph transformer on large-scale molecular data.
\newblock \emph{Advances in Neural Information Processing Systems}, 33: 12559--12571.

\bibitem[{Ross et~al.(2022)Ross, Belgodere, Chenthamarakshan, Padhi, Mroueh, and Das}]{ross2022large}
Ross, J.; Belgodere, B.; Chenthamarakshan, V.; Padhi, I.; Mroueh, Y.; and Das, P. 2022.
\newblock Large-scale chemical language representations capture molecular structure and properties.
\newblock \emph{Nature Machine Intelligence}, 4(12): 1256--1264.

\bibitem[{Seidl et~al.(2023)Seidl, Vall, Hochreiter, and Klambauer}]{seidl2023enhancing}
Seidl, P.; Vall, A.; Hochreiter, S.; and Klambauer, G. 2023.
\newblock Enhancing activity prediction models in drug discovery with the ability to understand human language.
\newblock \emph{arXiv preprint arXiv:2303.03363}.

\bibitem[{Song et~al.(2020)Song, Zheng, Niu, Fu, Lu, and Yang}]{song2020communicative}
Song, Y.; Zheng, S.; Niu, Z.; Fu, Z.-H.; Lu, Y.; and Yang, Y. 2020.
\newblock Communicative Representation Learning on Attributed Molecular Graphs.
\newblock In \emph{IJCAI}, volume 2020, 2831--2838.

\bibitem[{St{\"a}rk et~al.(2022)St{\"a}rk, Beaini, Corso, Tossou, Dallago, G{\"u}nnemann, and Li{\`o}}]{stark20223d}
St{\"a}rk, H.; Beaini, D.; Corso, G.; Tossou, P.; Dallago, C.; G{\"u}nnemann, S.; and Li{\`o}, P. 2022.
\newblock 3d infomax improves gnns for molecular property prediction.
\newblock In \emph{International Conference on Machine Learning}, 20479--20502. PMLR.

\bibitem[{Su et~al.(2022)Su, Du, Yang, Zhou, Li, Rao, Sun, Lu, and Wen}]{su2022molecular}
Su, B.; Du, D.; Yang, Z.; Zhou, Y.; Li, J.; Rao, A.; Sun, H.; Lu, Z.; and Wen, J.-R. 2022.
\newblock A molecular multimodal foundation model associating molecule graphs with natural language.
\newblock \emph{arXiv preprint arXiv:2209.05481}.

\bibitem[{Vaswani et~al.(2017)Vaswani, Shazeer, Parmar, Uszkoreit, Jones, Gomez, Kaiser, and Polosukhin}]{vaswani2017attention}
Vaswani, A.; Shazeer, N.; Parmar, N.; Uszkoreit, J.; Jones, L.; Gomez, A.~N.; Kaiser, {\L}.; and Polosukhin, I. 2017.
\newblock Attention is all you need.
\newblock \emph{Advances in neural information processing systems}, 30.

\bibitem[{Wang et~al.(2019)Wang, Guo, Wang, Sun, and Huang}]{wang2019smiles}
Wang, S.; Guo, Y.; Wang, Y.; Sun, H.; and Huang, J. 2019.
\newblock Smiles-bert: large scale unsupervised pre-training for molecular property prediction.
\newblock In \emph{Proceedings of the 10th ACM international conference on bioinformatics, computational biology and health informatics}, 429--436.

\bibitem[{Wang et~al.(2023)Wang, Bao, Dong, Bjorck, Peng, Liu, Aggarwal, Mohammed, Singhal, Som et~al.}]{wang2023image}
Wang, W.; Bao, H.; Dong, L.; Bjorck, J.; Peng, Z.; Liu, Q.; Aggarwal, K.; Mohammed, O.~K.; Singhal, S.; Som, S.; et~al. 2023.
\newblock Image as a Foreign Language: BEiT Pretraining for Vision and Vision-Language Tasks.
\newblock In \emph{Proceedings of the IEEE/CVF Conference on Computer Vision and Pattern Recognition}, 19175--19186.

\bibitem[{Wang et~al.(2022)Wang, Wang, Cao, and Barati~Farimani}]{wang2022molecular}
Wang, Y.; Wang, J.; Cao, Z.; and Barati~Farimani, A. 2022.
\newblock Molecular contrastive learning of representations via graph neural networks.
\newblock \emph{Nature Machine Intelligence}, 4(3): 279--287.

\bibitem[{Weininger(1988)}]{weininger1988smiles}
Weininger, D. 1988.
\newblock SMILES, a chemical language and information system. 1. Introduction to methodology and encoding rules.
\newblock \emph{Journal of chemical information and computer sciences}, 28(1): 31--36.

\bibitem[{Wiener(1947)}]{wiener1947structural}
Wiener, H. 1947.
\newblock Structural determination of paraffin boiling points.
\newblock \emph{Journal of the American chemical society}, 69(1): 17--20.

\bibitem[{Wishart et~al.(2018)Wishart, Feunang, Guo, Lo, Marcu, Grant, Sajed, Johnson, Li, Sayeeda et~al.}]{wishart2018drugbank}
Wishart, D.~S.; Feunang, Y.~D.; Guo, A.~C.; Lo, E.~J.; Marcu, A.; Grant, J.~R.; Sajed, T.; Johnson, D.; Li, C.; Sayeeda, Z.; et~al. 2018.
\newblock DrugBank 5.0: a major update to the DrugBank database for 2018.
\newblock \emph{Nucleic acids research}, 46(D1): D1074--D1082.

\bibitem[{Wu et~al.(2018)Wu, Ramsundar, Feinberg, Gomes, Geniesse, Pappu, Leswing, and Pande}]{wu2018moleculenet}
Wu, Z.; Ramsundar, B.; Feinberg, E.~N.; Gomes, J.; Geniesse, C.; Pappu, A.~S.; Leswing, K.; and Pande, V. 2018.
\newblock MoleculeNet: a benchmark for molecular machine learning.
\newblock \emph{Chemical science}, 9(2): 513--530.

\bibitem[{Xia et~al.(2022)Xia, Zhao, Hu, Gao, Tan, Liu, Li, and Li}]{xia2022mole}
Xia, J.; Zhao, C.; Hu, B.; Gao, Z.; Tan, C.; Liu, Y.; Li, S.; and Li, S.~Z. 2022.
\newblock Mole-bert: Rethinking pre-training graph neural networks for molecules.
\newblock In \emph{The Eleventh International Conference on Learning Representations}.

\bibitem[{Xia et~al.(2023)Xia, Zhu, Du, Liu, and Li}]{xia2023systematic}
Xia, J.; Zhu, Y.; Du, Y.; Liu, Y.; and Li, S. 2023.
\newblock A Systematic Survey of Chemical Pre-trained Models.
\newblock IJCAI.

\bibitem[{Yang et~al.(2019)Yang, Swanson, Jin, Coley, Eiden, Gao, Guzman-Perez, Hopper, Kelley, Mathea et~al.}]{yang2019analyzing}
Yang, K.; Swanson, K.; Jin, W.; Coley, C.; Eiden, P.; Gao, H.; Guzman-Perez, A.; Hopper, T.; Kelley, B.; Mathea, M.; et~al. 2019.
\newblock Analyzing learned molecular representations for property prediction.
\newblock \emph{Journal of chemical information and modeling}, 59(8): 3370--3388.

\bibitem[{Yasunaga, Leskovec, and Liang(2022)}]{yasunaga2022linkbert}
Yasunaga, M.; Leskovec, J.; and Liang, P. 2022.
\newblock Linkbert: Pretraining language models with document links.
\newblock \emph{arXiv preprint arXiv:2203.15827}.

\bibitem[{You et~al.(2020)You, Chen, Sui, Chen, Wang, and Shen}]{you2020graph}
You, Y.; Chen, T.; Sui, Y.; Chen, T.; Wang, Z.; and Shen, Y. 2020.
\newblock Graph contrastive learning with augmentations.
\newblock \emph{Advances in neural information processing systems}, 33: 5812--5823.

\bibitem[{Yu et~al.(2022)Yu, Wang, Vasudevan, Yeung, Seyedhosseini, and Wu}]{yu2022coca}
Yu, J.; Wang, Z.; Vasudevan, V.; Yeung, L.; Seyedhosseini, M.; and Wu, Y. 2022.
\newblock Coca: Contrastive captioners are image-text foundation models.
\newblock \emph{arXiv preprint arXiv:2205.01917}.

\bibitem[{Zaidi et~al.(2022)Zaidi, Schaarschmidt, Martens, Kim, Teh, Sanchez-Gonzalez, Battaglia, Pascanu, and Godwin}]{zaidi2022pre}
Zaidi, S.; Schaarschmidt, M.; Martens, J.; Kim, H.; Teh, Y.~W.; Sanchez-Gonzalez, A.; Battaglia, P.; Pascanu, R.; and Godwin, J. 2022.
\newblock Pre-training via denoising for molecular property prediction.
\newblock \emph{arXiv preprint arXiv:2206.00133}.

\bibitem[{Zeng et~al.(2022)Zeng, Yao, Liu, and Sun}]{zeng2022deep}
Zeng, Z.; Yao, Y.; Liu, Z.; and Sun, M. 2022.
\newblock A deep-learning system bridging molecule structure and biomedical text with comprehension comparable to human professionals.
\newblock \emph{Nature communications}, 13(1): 862.

\bibitem[{Zeng et~al.(2023)Zeng, Yin, Wang, Liu, Yang, Yao, Sun, Sun, Xie, and Liu}]{zeng2023interactive}
Zeng, Z.; Yin, B.; Wang, S.; Liu, J.; Yang, C.; Yao, H.; Sun, X.; Sun, M.; Xie, G.; and Liu, Z. 2023.
\newblock Interactive Molecular Discovery with Natural Language.
\newblock \emph{arXiv preprint arXiv:2306.11976}.

\bibitem[{Zhao et~al.(2023)Zhao, Liu, Ma, Xu, Fu, Deng, Kong, and Liu}]{zhao2023gimlet}
Zhao, H.; Liu, S.; Ma, C.; Xu, H.; Fu, J.; Deng, Z.-H.; Kong, L.; and Liu, Q. 2023.
\newblock GIMLET: A Unified Graph-Text Model for Instruction-Based Molecule Zero-Shot Learning.
\newblock \emph{bioRxiv}, 2023--05.

\bibitem[{Zhou et~al.(2023)Zhou, Gao, Ding, Zheng, Xu, Wei, Zhang, and Ke}]{zhou2023uni}
Zhou, G.; Gao, Z.; Ding, Q.; Zheng, H.; Xu, H.; Wei, Z.; Zhang, L.; and Ke, G. 2023.
\newblock Uni-Mol: a universal 3D molecular representation learning framework.

\end{thebibliography}

% \onecolumn
\appendix
\section{Detailed Experimental Results}

Due to space constraints in the main text, we omitted the standard deviations of the results from downstream experiments, providing only the average values from three trials. Tab. \ref{tab:complete-cls-results} displays the complete experimental results of our method and baselines on four classification tasks in MoleculeNet. Additionally, we employed $\Delta\mathrm{AP}$ to assess model performance, as utilized by \citet{seidl2023enhancing}. $\Delta\mathrm{AP}$ quantifies the overall effectiveness of a model by measuring the discrepancy between its ability to correctly identify classes (as captured by average precision) and the inherent frequency of those classes within the dataset (reflected by the base rate), calculated as follows:

\begin{equation}
    \Delta \mathrm{AP} = \frac{1}{N} \sum_{i=1}^{N} (\mathrm{AP}_i - \mathrm{BaseRate}_i)
\end{equation}
where, $N$ is the number of classes. 

To compute $\Delta\mathrm{AP}$, we first define the average precision ($\mathrm{AP}$) for class $i$:

\begin{equation}
    \mathrm{AP}_i = \sum_{k=1}^{M} (\mathrm{P}(k) \cdot \Delta r(k))
\end{equation}
where, $\mathrm{P}(k)$ represents the precision at the $k^{th}$ threshold, and $\Delta r(k)$ denotes the change in recall from the preceding threshold. $M$ stands for the total count of non-missing samples considered for the $i^{th}$ class.  $\mathrm{AP}$ is equivalent to the area under the precision-recall curve (AUPRC). Meanwhile, this value can be conveniently calculated using the function ``torchmetrics.functional.average\_precision()'' supported by TorchMetrics\footnote{\seqsplit{https://lightning.ai/docs/torchmetrics/stable/classification/average\_precision.html}}.

Furthermore, the base rate for class $i$, $\mathrm{BaseRate}_i$, is established by:
\begin{equation}
    \mathrm{BaseRate}_i = \frac{\sum_{j=1}^{M} y_{ij} \cdot \mathrm{mask}_{ij}}{\sum_{j=1}^{M} \mathrm{mask}_{ij}}
\end{equation}
where, $y_{ij}$ represents the binary label for the $j^{th}$ sample under class $i$. The term $\mathrm{mask}_{ij}$ indicates whether the value is valid. The complete code for calculating $\Delta\mathrm{AP}$ is available in the publicly shared code of this project under the class ``DeltaAveragePrecision''.

Tab. \ref{tab:complete-reg-results} presents the performance of models on regression tasks. Furthermore, we supplemented the performance of two recent molecular-text multimodal methods: TCT5 \cite{christofidellis2023unifying} and CLAMP \cite{seidl2023enhancing}.

\section{Data Overlap Analysis}
To demonstrate that the performance improvements brought by MolTailor were not caused by data leakage, we analyzed the overlap between the pre-training dataset MT-MTR and the downstream task datasets, as shown in Tab. \ref{tab:overlap-analysis}. The results indicate that there indeed was some data overlap.Following this, we delved deeper into our analysis by constructing a pre-training dataset MT-MTR$^\star$, which excluded the overlapping data. Subsequently, we trained MolTailor$^\star$ using the newly formed dataset, and the final experimental results are presented in Tab. \ref{tab:overlap-analysis}. 

The results reveal that there is no clear correlation between the degree of data overlap and the model's performance. Moreover, MolTailor$^\star$ still achieved excellent performance on regression tasks, though its performance on classification tasks was impaired. We believe this distinction underscores the dual impact of MT-MTR: it enhances the model's performance on regression tasks while conversely affecting its performance on classification tasks.

\section{Prompt Examples}
Due to policy adjustments at OpenAI, the prompts used to generate pre-training corpora and downstream task descriptions have become unavailable, as shown in Tab. \ref{tab:prompt}. Therefore, in this section, we only present data examples from MT-MTR and descriptions corresponding to downstream tasks. Tab. \ref{tab:mt-mtr-example} displays five data examples from MT-MTR, while Tab. \ref{tab:molnet-example} presents descriptions of eight tasks from the MoleculeNet dataset.

\section{Statistics of Label Counts in MT-MTR}
When constructing MT-MTR, it is theoretically randomized to select 5-10 properties from the 209 molecular properties provided by RDKit to generate task descriptions. Therefore, in theory, the number of the 209 properties in the regression labels of MT-MTR should be approximately equal. However, due to the high probability of properties starting with ``fr'', indicating the count of certain groups, being zero, we limited the selection probability of properties beginning with ``fr'' during the selection process, as detailed in the file ``03-generate-descriptions.py'' on GitHub. Additionally, we filtered out properties with excessively high absolute values or those that are NaN. 

\begin{landscape}
\begin{figure}[htbp]
    \centering
    \includegraphics[width=1\linewidth]{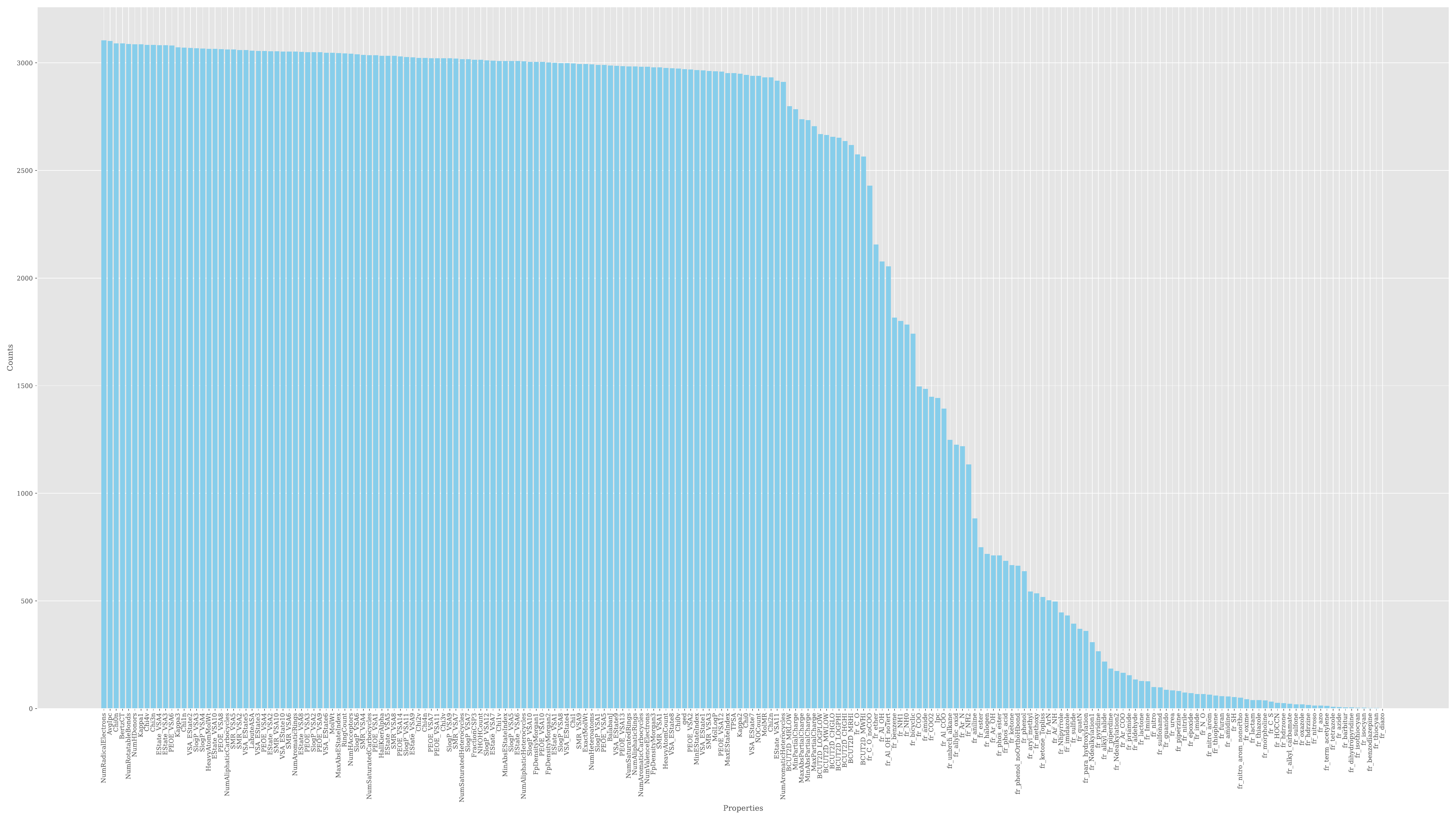}
    \caption{Statistical results of regression labels in MT-MTR. We counted the occurrence of each property within the regression labels, as provided by RDKit. The horizontal axis of the graph represents the name of the property, while the vertical axis indicates the number of occurrences of that property.}
    \label{fig:task_count}
\end{figure}
\end{landscape}

Following these filters, the statistical situation of the regression labels included in MT-MTR is shown in Fig. \ref{fig:task_count}. The statistical results reveal that, excluding properties beginning with 'fr', the occurrence probabilities of various properties are relatively similar, which aligns with expectations.

\renewcommand{\arraystretch}{0.8}
\begin{table*}[t]
    \centering
    \scalebox{0.9}{
      \begin{tabular}{l|cccccccc}
      \toprule
      \midrule
      \multicolumn{1}{c|}{\textbf{Dataset}} & \multicolumn{2}{c}{\textbf{BBBP}} & \multicolumn{2}{c}{\textbf{ClinTox}} & \multicolumn{2}{c}{\textbf{HIV}} & \multicolumn{2}{c}{\textbf{Tox21}} \\
      \midrule
      \#Molecules & \multicolumn{2}{c}{2039} & \multicolumn{2}{c}{1478} & \multicolumn{2}{c}{41127} & \multicolumn{2}{c}{7831} \\
      \#Split & \multicolumn{2}{c}{Scaffold} & \multicolumn{2}{c}{Random} & \multicolumn{2}{c}{Scaffold} & \multicolumn{2}{c}{Random} \\
      \#Tasks & \multicolumn{2}{c}{1} & \multicolumn{2}{c}{2} & \multicolumn{2}{c}{1} & \multicolumn{2}{c}{12} \\
      \midrule
      \#Metrics & ROC-AUC & $\Delta$AP   & ROC-AUC & $\Delta$AP   & ROC-AUC & $\Delta$AP   & ROC-AUC & $\Delta$AP \\
      \midrule
      Random & 48.38±3.43 & 0.83±2.51 & 56.01±4.79 & 2.09±0.48 & 49.54±1.27 & -0.20±0.10 & 51.11±1.86 & 0.98±0.32 \\
      RDKit-DP & 78.25±3.44 & 13.14±2.78 & 67.36±5.62 & 8.36±3.00 & 70.85±2.73 & 22.96±3.49 & 65.61±2.00 & 17.16±1.77 \\
      \midrule
      RDKit-FP & 87.65±1.70 & 20.40±0.95 & 57.13±2.68 & 3.00±1.54 & 78.66±0.88 & 34.32±3.35 & 76.14±0.96 & 32.39±0.89 \\
      MACCS-FP & 81.64±1.97 & 17.22±1.73 & 83.05±13.13 & 18.53±9.54 & 77.53±1.29 & 23.09±2.15 & 77.27±0.71 & 27.19±1.04 \\
      Morgan-FP & 82.73±4.46 & 17.14±2.86 & 72.61±5.94 & 12.92±3.04 & 82.65±0.26 & 41.09±3.17 & 75.29±1.49 & 33.77±0.84 \\
      \midrule
      Grover & 79.83±1.48 & 15.87±1.91 & 87.75±8.38 & 20.60±9.74 & 77.47±2.70 & 25.42±3.66 & 79.61±0.37 & 25.55±3.67 \\
      MolCLR & 81.27±4.57 & 16.70±2.32 & 78.15±2.29 & 11.03±1.44 & 71.48±5.56 & 13.00±3.73 & 75.61±1.62 & 22.18±1.06 \\
      Mole-BERT & 82.70±2.72 & 17.18±2.48 & 81.82±4.62 & 15.72±1.65 & 79.35±1.84 & 25.40±4.27 & 84.20±0.71 & 35.77±0.70 \\
      Uni-Mol & 79.52±1.97 & 16.35±0.83 & 88.65±2.09 & 23.75±4.32 & 74.18±3.05 & 22.54±5.90 & 78.08±1.60 & 26.11±1.48 \\
      \midrule
      BioLinkBERT & 83.81±3.54 & 17.86±1.65 & 87.75±6.30 & 21.29±5.09 & 71.24±5.20 & 22.36±7.71 & 73.81±1.20 & 18.69±1.12 \\
      PubMedBERT & \textbf{89.10±1.73} & \textbf{20.74±0.28} & 84.29±6.79 & 18.91±3.97 & 72.30±4.91 & 20.44±6.26 & 73.77±1.17 & 19.55±1.11 \\
      ChemBERTa-2 & 84.70±2.17 & 18.39±2.53 & 84.21±0.61 & 13.41±3.36 & 78.88±1.19 & 32.40±6.44 & 80.75±1.74 & 30.90±1.40 \\
      CHEM-BERT & 84.10±2.25 & 16.98±2.77 & \textbf{93.80±1.11} & 26.82±2.56 & 76.99±3.03 & 26.96±5.93 & 80.54±1.04 & 28.75±0.45 \\
      \midrule
      KCL   & 76.86±1.28 & 12.99±1.88 & 60.80±13.17 & 4.99±3.55 & 68.48±3.89 & 11.04±4.04 & 74.98±2.08 & 18.34±2.50 \\
      KV-PLM & 86.36±1.38 & 19.24±1.35 & 81.20±8.53 & 15.05±6.80 & 73.52±3.94 & 21.52±6.17 & 74.62±1.22 & 21.55±3.10 \\
      MoMu-ME  & 80.41±2.91 & 16.43±1.95 & 67.99±8.41 & 6.44±2.16 & 71.91±1.95 & 12.16±2.63 & 74.75±2.26 & 19.72±1.08 \\
      MoMu-TE & 82.24±2.35 & 17.57±1.29 & 81.94±3.01 & 12.54±2.53 & 67.88±3.25 & 13.37±3.15 & 73.07±1.16 & 20.32±0.22 \\
      MolT5 & 86.77±2.68 & 19.62±1.77 & 89.65±7.45 & 25.07±11.90 & 77.14±4.89 & 27.55±9.37 & 80.66±1.28 & 31.69±1.83 \\
      TCT5  & 85.58±2.51 & 18.51±18.51 & 88.73±5.55 & 22.77±10.73 & 79.33±1.22 & 28.30±4.84 & 81.72±1.61 & 32.44±1.80 \\
      CLAMP & 83.08±0.80 & 16.40±0.64 & 83.02±2.85 & 15.98±2.00 & \textbf{93.38±0.79} & \textbf{60.75±2.55} & \textbf{85.14±1.67} & \textbf{43.67±2.02} \\
      \midrule
      MolTailor & 81.15±2.15 & 15.94±1.84 & 92.37±4.45 & \underline{\textbf{29.58±2.70}} & \underline{77.42±2.33} & 26.23±4.87 & \underline{80.67±1.36} & \underline{30.57±0.81} \\
      MolTailor$^*$ & 84.65±2.12 & 18.31±1.01 & \underline{85.95±6.67} & \underline{19.88±7.53} & 76.42±2.63 & 24.77±4.50 & 80.32±1.59 & \underline{31.10±1.72} \\
      MolTailor$^{**}$ & 83.08±1.09 & 17.43±0.30 & 83.65±5.23 & \underline{20.11±5.96} & 76.01±1.30 & 26.55±2.59 & \underline{80.90±1.91} & \underline{31.38±1.02} \\
      \midrule
      \bottomrule
      \end{tabular}%
    }
    \caption{Experimental results on classification tasks. All experiments were conducted under the linear probe setting, and for each method across each task, the learning rate was searched ten times using Optuna. We conducted repeated experiments using three non-specifically set random seeds: 1236, 1237, 1238, and reported the mean and standard deviation of these three trials. TCT5 \cite{christofidellis2023unifying} and CLAMP \cite{seidl2023enhancing} are additionally included as two new multimodal methods. Additionally, we employed $\Delta$AP to assess model performance, as utilized by \citet{seidl2023enhancing}. MolTailor denotes the use of PubMedBERT and CHEM-BERT as the backbone, MolTailor$^*$ represents the use of PubMedBERT and ChemBERTa as the backbone, and MolTailor$^{**}$ indicates the use of BioLinkBERT and CHEM-BERT as the backbone. Bold results indicate the best outcomes, while underlined results signify that MolTailor shows an improvement over the molecular backbone used.}
    \label{tab:complete-cls-results}%
\end{table*}%
\renewcommand{\arraystretch}{1}

% Table generated by Excel2LaTeX from sheet 'Sheet5'
\renewcommand{\arraystretch}{0.8}
\begin{table*}[t]
    \centering
    \scalebox{0.9}{
      \begin{tabular}{l|cccc|r}
      \toprule
      \midrule
      \multicolumn{1}{c|}{\textbf{Dataset}} & \textbf{ESOL} & \textbf{FreeSolv} & \textbf{Lip} & \textbf{QM8} & \multicolumn{1}{c}{\multirow{5}[6]{*}{\textbf{Params}}} \\
  \cmidrule{1-5}    \#Molecules & 1128  & 642   & 4200  & 21786 &  \\
      \#Split & Random & Random & Random & Random &  \\
      \#Tasks & 1     & 1     & 2     & 16    &  \\
  \cmidrule{1-5}    \#Metrics & RMSE  & RMSE  & RMSE  & MAE   &  \\
      \midrule
      Random & 3.3358±0.0904 & 5.4831±0.6031 & 1.3813±0.0170 & 0.0320±0.0004 & - \\
      RDKit-DP & 4.8940±4.9635 & 2.8068±0.9990 & 0.9963±0.0448 & 0.0202±0.0006 & - \\
      \midrule
      RDKit-FP & 1.0830±0.1084 & 2.0725±0.1017 & 0.9007±0.0248 & \textbf{0.0181±0.0005} & - \\
      MACCS-FP & 1.0833±0.1146 & 1.9086±0.2425 & 0.9886±0.0546 & 0.0196±0.0005 & - \\
      Morgan-FP & 1.2413±0.1281 & 2.1896±0.4250 & 0.8196±0.0252 & 0.0200±0.0006 & - \\
      \midrule
      Grover & 0.8977±0.0594 & 1.9041±0.1993 & 0.8301±0.0414 & 0.0184±0.0007 & 107M \\
      MolCLR & 1.3421±0.0578 & 3.0436±0.7911 & 1.0448±0.0288 & 0.0219±0.0007 & 2M \\
      Mole-BERT & 1.1379±0.0362 & 2.3626±0.1184 & 0.8316±0.0320 & 0.0221±0.0007 & 2M \\
      Uni-Mol & 1.0509±0.0801 & 2.6913±0.4001 & 1.0363±0.0617 & 0.0219±0.0005 & 48M \\
      \midrule
      BioLinkBERT & 1.1739±0.0472 & 3.1350±0.1649 & 1.0589±0.0506 & 0.0234±0.0008 & 108M \\
      PubMedBERT & 1.0715±0.0095 & 2.5999±0.4324 & 1.0851±0.0491 & 0.0232±0.0007 & 108M \\
      ChemBERTa-2 & 0.8103±0.0530 & 1.8439±0.5930 & 0.7948±0.0209 & 0.0191±0.0008 & 3M \\
      CHEM-BERT & 0.7973±0.0462 & 2.0214±0.5824 & 0.8571±0.0381 & 0.0215±0.0011 & 51M \\
      \midrule
      KCL   & 0.8728±0.0524 & 2.7615±0.2813 & 0.9868±0.0368 & 0.0225±0.0007 & 1M \\
      KV-PLM & 1.1785±0.1242 & 2.8840±0.3591 & 1.1004±0.0242 & 0.0233±0.0009 & 109M \\
      MoMu-ME  & 1.4135±0.0504 & 2.3229±0.0986 & 0.9835±0.0600 & 0.0222±0.0007 & 2M \\
      MoMu-TE & 1.2562±0.1029 & 3.1480±0.4755 & 1.0885±0.0116 & 0.0250±0.0006 & 109M \\
      MolT5 & 0.7867±0.0464 & 2.0978±0.2920 & 0.8760±0.8760 & 0.0204±0.0007 & 110M \\
      TCT5  & 0.8194±0.0356 & 1.8784±0.2288 & 0.8269±0.0377 & 0.0189±0.0006 & 110M \\
      CLAMP & 0.8012±0.0651 & 2.1808±0.0604 & 0.7978±0.0499 & 0.0205±0.0007 & 56M \\
      \midrule
      MolTailor & \underline{0.7234±0.0495} & \underline{1.7881±0.3262} & \underline{0.8107±0.0160} & \underline{0.0196±0.0005} & 161M \\
      MolTailor$^*$ & \underline{\textbf{0.7128±0.0459}} & \underline{\textbf{1.7826±0.3342}} & \underline{\textbf{0.7848±0.0122}} & \underline{0.0185±0.0005} & 112M \\
      MolTailor$^{**}$ & \underline{0.7329±0.0331} & \underline{1.7845±0.4055} & 0.8094±0.0034 & \underline{0.0184±0.0006} & 112M \\
      \midrule
      \bottomrule
      \end{tabular}%
    }
    \caption{Experimental results on regression tasks. The experimental setup is identical to classification tasks, as referenced in Tab. \ref{tab:complete-cls-results}. TCT5 \cite{christofidellis2023unifying} and CLAMP \cite{seidl2023enhancing} are additionally included as two new multimodal methods. MolTailor denotes the use of PubMedBERT and CHEM-BERT as the backbone, MolTailor$^*$ represents the use of PubMedBERT and ChemBERTa as the backbone, and MolTailor$^{**}$ indicates the use of BioLinkBERT and CHEM-BERT as the backbone. Bold results indicate the best outcomes, while underlined results signify that MolTailor shows an improvement over the molecular backbone used.}
    \label{tab:complete-reg-results}%
\end{table*}%
\renewcommand{\arraystretch}{1}

% Table generated by Excel2LaTeX from sheet 'Sheet6'
\begin{table*}[t]
  \centering

    \begin{tabular}{c|c|c|cccccc}
    \toprule
    \midrule
    \multicolumn{3}{c|}{\textbf{Dataset}} & \multicolumn{3}{c}{\textbf{Overlapping Ratio}} & \textbf{CHEM-BERT} & \textbf{MolTailor} & \textbf{MolTailor$^\star$} \\
    \midrule
    \multirow{5}[4]{*}{\begin{sideways}\textbf{Classification}\end{sideways}} & \multirow{5}[4]{*}{\begin{sideways}ROC-AUC\end{sideways}} & BBBP  & 1975  & 909   & 46.03\% & \textbf{84.10±2.25} & 81.15±2.15 & \underline{83.04±0.62} \\
          &       & ClinTox & 1459  & 766   & 52.50\% & \textbf{93.80±1.11} & \underline{92.37±4.45} & 91.89±2.05 \\
          &       & HIV   & 41127 & 618   & 1.50\% & \underline{76.99±3.03} & \textbf{77.42±2.33} & 76.09±2.94 \\
          &       & Tox21 & 7831  & 3937  & 50.27\% & \underline{80.54±1.04} & \textbf{80.67±1.36} & 80.25±2.13 \\
\cmidrule{3-9}          &       & Average & -     & -     & -     & \textbf{83.86 } & \underline{82.90}  & 82.82  \\
    \midrule
    \multirow{5}[4]{*}{\begin{sideways}\textbf{Regression}\end{sideways}} & \multirow{5}[4]{*}{\begin{sideways}RMSE / MAE\end{sideways}} & ESOL  & 1117  & 651   & 58.28\% & 0.7973±0.0462 & \textbf{0.7234±0.0495} & \underline{0.7602±0.0619} \\
          &       & FreeSolv & 642   & 407   & 63.40\% & 2.0214±0.5824 & \underline{1.7881±0.3262} & \textbf{1.7097±0.2594} \\
          &       & Lip & 4200  & 629   & 14.98\% & 0.8571±0.0381 & \underline{0.8107±0.0160} & \textbf{0.8095±0.0364} \\
          &       & QM8   & 21766 & 423   & 1.94\% & 0.0215±0.0011 & \textbf{0.0196±0.0005} & \underline{0.0197±0.0008} \\
\cmidrule{3-9}          &       & Average & -     & -     & -     & 0.9243  & \underline{0.8355}  & \textbf{0.8248} \\
    \midrule
    \bottomrule
    \end{tabular}%
    \caption{Data overlap analysis results. We examined the overlap of SMILES between MT-MTR and eight tasks from MoleculeNet, with the analysis results presented in the left half of the table. In the three columns under ``Overlapping Ratio'', the first column indicates the number of molecules contained in the dataset, the second column shows the number of molecules that overlap with those in MT-MTR, and the third column represents the proportion of overlap. It is important to note that due to our preprocessing of SMILES which includes normalization and deduplication, the number of molecules in the table differs from the counts originally reported. On the right, we display the performance changes of the models after removing data overlap, where CHEM-BERT denotes the performance of the molecular backbone used, MolTailor refers to the model trained on the original MT-MTR dataset, and MolTailor$^\star$ refers to the model trained on MT-MTR$^\star$, the newly constructed dataset with overlapping data removed.}
  \label{tab:overlap-analysis}%
\end{table*}%

\clearpage

\onecolumn
% \begin{landscape}
% \renewcommand{\arraystretch}{0.8}
\begin{longtable}{c|p{45em}}
    \toprule
    \midrule
            & \multicolumn{1}{l}{\textbf{MT-MTR Examples}} \\
    \midrule
    \endfirsthead

    \multicolumn{2}{c}%
    {\tablename\ \thetable\ -- \textit{Continued from previous page}} \\
    \toprule
    & \textbf{MT-MTR Examples} \\
    \midrule
    \endhead
    1     & SMILES:\newline{}\texttt{\seqsplit{COc1cccc2c1C(=O)c1c(O)c3c(c(O)c1C2=O)C[C@@](O)(C(=O)CO)C[C@@H]3O[C@H]1C[C@H]([NH3+])[C@@H](O)[C@H](C)O1}}\newline{}\newline{}Task Descriptions:\newline{}\texttt{Based on the analysis of the chemical compound, the following properties are important in solving the task:\newline{}1. NumSaturatedHeterocycles: This descriptor provides information about the number of saturated heterocycles present in the compound. It helps in understanding the presence of specific ring structures that can be relevant for the task.\newline{}2. Chi4n: This descriptor measures the 4th-order molecular connectivity index. It indicates the electronic and topological properties of the compound. It can be useful in identifying the compound's reactivity and structural complexity.\newline{}3. VSA\_EState5: This descriptor represents the potential electrophilicity of the compound. It provides insight into the compound's ability to act as an electron acceptor in chemical reactions.\newline{}4. VSA\_EState7: This descriptor is associated with the electrophilic properties of the compound. It helps in understanding the compound's electron-rich regions and its ability to act as an electron donor in reactions.\newline{}5. NumAromaticRings: This descriptor indicates the number of aromatic rings present in the compound. It helps in identifying the presence of specific ring structures that contribute to the compound's aromaticity.\newline{}By considering these properties, we can gain valuable insights into the compound's structure, reactivity, and presence of important functional groups or ring systems. These insights can aid in understanding the compound's behavior and its potential applications in various chemical tasks.} \\
    \midrule
    2     & SMILES:\newline{}\texttt{\seqsplit{C[NH+]1[C@H]2C[C@H](OC(=O)C(CO)c3ccccc3)C[C@@H]1[C@@H](O)C2}}\newline{}\newline{}Task Description:\newline{}\texttt{Based on the given descriptors, the following analysis can be made:\newline{}1. SlogP\_VSA2: This descriptor represents the hydrophobicity of the compound. A higher value indicates a higher tendency for the compound to dissolve in lipid-like environments.\newline{}2. fr\_C\_O: This descriptor counts the number of occurrences of a carbon-oxygen bond in the compound. It provides information about the presence of carbonyl groups, which are important in various chemical reactions.\newline{}3. TPSA: The Topological Polar Surface Area (TPSA) characterizes the polarity of a compound. A larger TPSA value suggests a higher degree of intermolecular interactions, which can affect solubility and permeability.\newline{}4. Chi3n: This descriptor calculates the third component of the molecular connectivity index, which is related to three-bond or three-connectivity information. It contributes to understanding the molecular architecture and reactivity.\newline{}5. NumHAcceptors: This descriptor counts the number of hydrogen bond acceptor groups in the compound. It represents the compound's ability to form hydrogen bonds with other molecules, influencing solvation and chemical interactions.\newline{}6. NumSaturatedCarbocycles: This descriptor counts the number of saturated carbocycles present in the compound. It provides information about the compound's rigidity and structural characteristics.\newline{}7. fr\_Al\_OH: This descriptor counts the number of hydroxyl groups attached to an alkene or aromatic ring. It is relevant in the context of reactions involving alcohols and aromatic compounds.\newline{}8. PEOE\_VSA8: This descriptor represents the partial equalization of orbital electronegativity (PEOE) potential energy (VSA) associated with the eighth bin. It contributes to understanding the compound's electronic characteristics and intermolecular interactions.\newline{}9. BCUT2D\_MRHI: This descriptor calculates the maximum value of the Burden Cluster-based Van der Waals atomic surface area. It provides information about the surface properties of the compound, influencing its solubility and reactivity.\newline{}Overall, analyzing these descriptors can provide valuable insights into the hydrophobicity, reactivity, polarity, hydrogen bonding potential, structural characteristics, and electronic properties of a chemical compound.} \\
    \midrule
    3     & SMILES:\newline{}\texttt{\seqsplit{O=C([O-])CC([O-])(CC(=O)[O-])C(=O)[O-]}}\newline{}\newline{}Task Description:\newline{}\texttt{Based on the given descriptors, the following analysis results have been observed for the chemical compound:\newline{}1. BCUT2D\_MWLOW: This descriptor calculates the molecular weight of the compound. It indicates the lower end of the molecular weight range. This information helps determine the compound's size and potential for various chemical reactions.\newline{}2. BalabanJ: This descriptor calculates the Balaban connectivity index, which measures the compound's molecular complexity and how connected its atoms are. Higher values indicate more complex and interconnected structures.\newline{}3. Chi1v: This descriptor calculates the chi connectivity index, which characterizes the topological features of a molecule. It provides information about the compound's branching and connectivity patterns.\newline{}4. NumAliphaticCarbocycles: This descriptor counts the number of aliphatic carbocycles present in the compound. It helps identify the presence of cyclic structures in the molecule, which can impact its stability and reactivity.\newline{}5. SlogP\_VSA12: This descriptor quantifies the topological polar surface area (TPSA) contribution to the compound's logarithm of the octanol/water partition coefficient (SlogP). It provides information about the compound's hydrophobic/hydrophilic nature.\newline{}6. PEOE\_VSA13: This descriptor represents the partial charges on the atoms of a molecule, specifically for peripheral oxygen atoms. It characterizes the compound's electronic distribution and potential interactions with other molecules.\newline{}By analyzing these properties, we can gain insights into the compound's molecular weight, complexity, branching patterns, presence of cyclic structures, hydrophobicity/hydrophilicity, and electronic distribution. These findings are essential for understanding the compound's behavior, stability, and potential reactions in various chemical processes.} \\
    \midrule
    4     & SMILES:\newline{}\texttt{\seqsplit{C1=Cc2cc3cc(-c4c5nc(cc6ccc(cc7nc(cc8ccc4[nH]8)C=C7)[nH]6)C=C5)c(cc4nc(cc5ccc(cc1n2)[nH]5)C=C4)[nH]3}}\newline{}\newline{}Task Description:\newline{}\texttt{Based on the given descriptors, the analysis of the chemical task indicates that several properties of the compound are influential in solving the task. \newline{}The EState\_VSA10 value represents the electronic state of the compound, specifically the tendency to donate or accept electrons. A higher value suggests a higher ability to interact with other molecules through electron transfer.\newline{}EState\_VSA7 measures the electronic state in terms of the compound's charge distribution. A higher value indicates a greater emphasis on electron density in specific regions, which may affect the compound's reactivity.\newline{}PEOE\_VSA13 and PEOE\_VSA14 are related to the partial charges of the compound, specifically in terms of polarizability. These values help understand the nature of charge distribution and electron density in the molecule.\newline{}EState\_VSA11 represents the electronic state, particularly the compound's ability to act as an electron donor. A higher value implies a stronger electron-donating capability.\newline{}SlogP\_VSA8 measures the lipophilicity or solubility of the compound. A higher value suggests a greater affinity for lipid-based solvents or hydrophobic environments.\newline{}Chi0n denotes the topological index, specifically related to electronegativity. It helps understand the tendency of the compound to attract electrons.\newline{}To effectively solve the chemical task, consideration must be given to these descriptors, which provide crucial insights into the compound's electronic and physical properties. By understanding these characteristics, a comprehensive analysis of the compound's behavior and potential reactions can be achieved.} \\
    \midrule
    5     & SMILES:\newline{}\texttt{\seqsplit{N[C@@H](CO)C(=O)N[C@@H](CCC(=O)O)C(=O)O}}\newline{}\newline{}Task Description:\newline{}\texttt{Based on the given descriptors, the analysis of the chemical compound is as follows:\newline{}1. FpDensityMorgan2: This descriptor measures the density of functional groups in the compound. It can provide information about the complexity and reactivity of the compound.\newline{}2. NumHeteroatoms: This descriptor counts the number of heteroatoms (atoms other than carbon and hydrogen) present in the compound. It helps in understanding the compound's potential for various reactions and interactions.\newline{}3. SMR\_VSA9: This descriptor is related to the surface area of the compound and provides information about its size and shape. It can be useful in predicting the compound's solubility and stability.\newline{}4. Chi1v: This descriptor represents the first molecular connectivity index, which measures the topological structure of the compound. It can help in understanding the compound's chemical behavior, reactivity, and properties.\newline{}5. BCUT2D\_CHGLO: This descriptor represents the Burden modified characteristic topological index, which measures the charge distribution in the compound. It can provide insights into the compound's polarity and electrostatic interactions.\newline{}6. SlogP\_VSA3: This descriptor is related to the compound's hydrophobicity and lipophilicity. It can provide information about its ability to dissolve in lipid-based environments.\newline{}By considering these descriptors, we can gain valuable insights into the structural and chemical properties of the compound, which can help in identifying its potential uses, reactions, and behavior.} \\
    \midrule
    \bottomrule
    \caption{Five data examples from MT-MTR.}
    \label{tab:mt-mtr-example}%
\end{longtable}%
% \renewcommand{\arraystretch}{1}
% \end{landscape}
\twocolumn

% Table generated by Excel2LaTeX from sheet 'Sheet8'
\onecolumn
\renewcommand{\arraystretch}{0.8}

\begin{longtable}{c|p{44.585em}}
    \toprule
    \midrule
    \textbf{Tasks} & \multicolumn{1}{l}{\textbf{Descriptions}} \\
    \midrule
    \endfirsthead

    \multicolumn{2}{c}%
    {\tablename\ \thetable\ -- \textit{Continued from previous page}} \\
    \toprule
    \textbf{Tasks} & \multicolumn{1}{l}{\textbf{Descriptions}} \\
    \midrule
    \endhead
    BBBP  & \texttt{Analyzing the properties related to this task, the following descriptors might be helpful:\newline{}1. MolLogP: This descriptor calculates the lipophilicity of the compound, which can influence its ability to penetrate the blood-brain barrier.\newline{}2. NumHDonors and NumHAcceptors: These descriptors count the number of hydrogen bond donors and acceptors, respectively, and can provide insights into the compound's solubility and permeability.\newline{}3. TPSA: The topological polar surface area can indicate the compound's polarity, affecting its interaction with the barrier.\newline{}4. fr\_halogen: This descriptor identifies the presence of halogen atoms, which might influence the compound's reactivity and ability to cross the barrier.\newline{}5. RingCount: The number of rings in a compound can affect its shape and size, influencing its ability to penetrate the barrier.\newline{}6. MolWt: Molecular weight can provide information about the size of the compound, which might be a factor in its ability to cross the blood-brain barrier.\newline{}7. FractionCSP3: This descriptor calculates the fraction of sp3 hybridized carbons, which can provide insights into the compound's three-dimensional structure and flexibility.\newline{}In summary, these descriptors can provide valuable insights into a compound's size, shape, polarity, and lipophilicity.} \\
    \midrule
    ClinTox & \texttt{The following descriptors could be instrumental in solving this task:\newline{}1. MolWt: Molecular weight can influence the compound's bioavailability and potential toxicity.\newline{}2. NumHAcceptors and NumHDonors: These descriptors provide information about hydrogen bonding, which can affect how a compound interacts with biological systems.\newline{}3. TPSA: Topological polar surface area can indicate how a compound might interact with membranes and proteins, influencing its toxicity.\newline{}4. fr\_halogen: The presence of halogen atoms can affect the compound's reactivity and potential toxicity.\newline{}5. fr\_nitro: This descriptor identifies nitro groups, often associated with toxic effects.\newline{}6. MolLogP: Lipophilicity, measured by this descriptor, can influence how a compound is distributed in the body, affecting its toxicity.\newline{}7. NumAromaticRings: Aromatic rings can play a role in the compound's stability and reactivity, influencing its toxic effects.\newline{}8. fr\_urea, fr\_sulfide, fr\_ketone: These functional group descriptors can provide insights into specific chemical reactivity that might be associated with toxicity.\newline{}In summary, these descriptors encompass various aspects of a compound's size, shape, polarity, lipophilicity, and functional groups.} \\
    \midrule
    HIV   & \texttt{Analyzing the task, the following descriptors could be particularly helpful in solving the task:\newline{}1. MolWt: Molecular weight can be indicative of the compound's ability to penetrate the HIV viral envelope.\newline{}2. NumHAcceptors and NumHDonors: These descriptors can provide insights into the hydrogen bonding capabilities, which might influence binding with HIV proteins.\newline{}3. MolLogP: This descriptor represents the compound's lipophilicity, which can affect its interaction with the viral membrane.\newline{}4. fr\_alkyl\_halide, fr\_amide, fr\_nitro: These functional group descriptors may indicate specific reactivity patterns with HIV enzymes.\newline{}5. RingCount: The number of rings in the compound may influence its conformation and ability to bind to HIV targets.\newline{}6. FractionCSP3: This descriptor calculates the fraction of sp3 hybridized carbons, providing insights into the compound's stereochemistry, which can be crucial for binding specificity.\newline{}In summary, these descriptors encompass various aspects of molecular structure and reactivity} \\
    \midrule
    Tox21 & \texttt{Analyzing the properties related to this task, the following descriptors could be helpful:\newline{}1. NumHAcceptors: This descriptor counts the number of hydrogen bond acceptors, which can indicate potential biological interactions and toxicity.\newline{}2. NumRadicalElectrons: Counting unpaired electrons can provide insights into the compound's reactivity, which may be linked to toxicity.\newline{}3. MolLogP: This descriptor calculates the octanol-water partition coefficient, reflecting the compound's hydrophobicity, which can influence its biological activity and potential toxicity.\newline{}4. fr\_halogen: Identifying the presence of halogen functional groups can be crucial, as certain halogenated compounds are known to be toxic.\newline{}5. fr\_nitro: This descriptor identifies nitro groups, which can be associated with toxic effects in some compounds.\newline{}6. TPSA: The topological polar surface area can provide information about the compound's ability to cross biological membranes, influencing its toxicity.\newline{}7. RingCount: The number of rings in a compound can affect its structural stability and interactions with biological targets, potentially impacting toxicity.\newline{}In summary, these descriptors can provide valuable insights into the compound's structure, reactivity, and interactions} \\
    \midrule
    ESOL  & \texttt{Analyzing this task, the following descriptors could be helpful in solving it:\newline{}1. MolWt: Molecular weight can influence solubility, as smaller molecules generally dissolve more easily.\newline{}2. TPSA: The topological polar surface area reflects the compound's polarity, affecting its interaction with polar solvents like water.\newline{}3. NumHDonors and NumHAcceptors: These descriptors count hydrogen bond donors and acceptors, respectively, and can provide insights into the compound's ability to form hydrogen bonds with water.\newline{}4. MolLogP: This descriptor measures the compound's lipophilicity, which can indicate its solubility in water versus organic solvents.\newline{}5. FractionCSP3: The fraction of sp3 hybridized carbons can provide information about the compound's three-dimensional structure and its potential solubility.\newline{}6. fr\_ether, fr\_aldehyde, fr\_ketone: These descriptors identify specific functional groups that may influence solubility.\newline{}In summary, these descriptors can offer valuable insights into a compound's solubility in water, considering factors like molecular weight, polarity, hydrogen bonding capability, lipophilicity, and specific functional groups.} \\
    \midrule
    FreeSolv & \texttt{Analyzing the properties related to this task, the following descriptors may be helpful:\newline{}1. MolWt: Molecular weight can influence solubility and interaction with water molecules.\newline{}2. TPSA: The topological polar surface area can provide insights into the polarity of the compound, affecting its solvation.\newline{}3. NumHDonors and NumHAcceptors: These descriptors count the number of hydrogen bond donors and acceptors, respectively, which can influence the compound's interaction with water.\newline{}4. MolLogP: This descriptor calculates the octanol-water partition coefficient, reflecting the compound's hydrophobicity or hydrophilicity.\newline{}5. FractionCSP3: The fraction of sp3 hybridized carbons can provide information about the compound's stereochemistry and potential interactions with water.\newline{}6. fr\_ether, fr\_aldehyde, fr\_ketone: These descriptors identify specific functional groups that may influence solvation behavior.\newline{}In summary, these descriptors can provide valuable insights into the compound's electronic distribution, polarity, hydrogen bonding capability, and specific functional group presence} \\
    \midrule
    Lip   & \texttt{Analyzing this task, the following descriptors could be helpful:\newline{}1. MolLogP: This descriptor calculates the logarithm of the compound's partition coefficient between n-octanol and water, providing direct insights into the compound's lipophilicity.\newline{}2. SlogP\_VSA1 to SlogP\_VSA12: These descriptors represent the surface area of a molecule contributing to its solubility in octanol, which can be correlated with lipophilicity.\newline{}3. NumHDonors and NumHAcceptors: These descriptors count the number of hydrogen bond donors and acceptors, respectively, which can influence the compound's interaction with lipid bilayers.\newline{}4. fr\_alkyl\_halide, fr\_ether, fr\_ester: These descriptors identify specific functional groups that may affect the compound's lipophilic behavior.\newline{}5. FractionCSP3: This descriptor calculates the fraction of carbon atoms that are sp3 hybridized, providing insights into the compound's stereochemistry, which may influence lipophilicity.\newline{}In summary, these descriptors can provide valuable information about the compound's lipophilic nature and interactions with lipid environments} \\
    \midrule
    QM8   & \texttt{Several descriptors can be identified as potentially significant for understanding and solving the task:\newline{}1. MaxEStateIndex \& MinEStateIndex: These descriptors calculate the maximum and minimum E-State indices of atoms in a compound, respectively. They can provide insights into the electronic state and distribution of charge within the molecule.\newline{}2. MolWt \& ExactMolWt: Representing the molecular weight and exact molecular weight, these descriptors can give information about the size and composition of the molecule.\newline{}3. NumValenceElectrons: This descriptor counts the number of valence electrons in a compound, which can be crucial for understanding its bonding and reactivity.\newline{}4. TPSA: The topological polar surface area can provide insights into the molecule's solubility and permeability.\newline{}5. EState\_VSA1 to EState\_VSA11: These descriptors calculate the E-State values for various atom types and can provide information about the electronic distribution and interactions within the compound.\newline{}6. VSA\_EState1 to VSA\_EState9: Representing the van der Waals surface area by estimating the electronic energy of atoms in a molecule, these descriptors can provide insights into the electronic distribution and interactions within the compound.\newline{}7. FractionCSP3: This descriptor calculates the fraction of carbon atoms in the compound that are sp3 hybridized, providing insights into the compound's stereochemistry.\newline{}8. NumAromaticRings \& NumAliphaticRings: These descriptors count the number of aromatic and aliphatic rings, respectively, in a compound, indicating the presence of cyclic structures.\newline{}9. MolLogP: This descriptor provides the predicted octanol-water partition coefficient, which can be crucial for understanding the compound's solubility and bioavailability.\newline{}10. fr\_ketone \& fr\_sulfide: Identifying the presence of ketone and sulfide functional groups, these descriptors can help in determining the compound's chemical reactivity and potential interactions.\newline{}In summary, the above-listed descriptors can offer valuable insights into the electronic distribution, reactivity, and other properties of the compounds in the task. Understanding these properties can be instrumental in solving the task at hand.} \\
    \midrule
    \bottomrule
    \caption{Descriptions for the datasets from MoleculeNet.}
    \label{tab:molnet-example}%
\end{longtable}%
\renewcommand{\arraystretch}{1}
\twocolumn

\onecolumn
\twocolumn

\end{document}